\documentclass[nohyperref]{article}

\usepackage{hyperref}

\usepackage[accepted]{icml2022}
\usepackage{microtype}
\usepackage{subfigure}
\usepackage{booktabs} % for professional tables
\usepackage{lscape}
\usepackage{rotating}
\usepackage[ruled, vlined, linesnumbered]{algorithm2e}
\usepackage[utf8]{inputenc}
\usepackage{microtype}
\usepackage{graphicx}
\usepackage{booktabs} % for professional tables
\usepackage{float}
\usepackage{url}            % simple URL typesetting
\usepackage{nicefrac}       % compact symbols for 1/2, etc.
\usepackage{xspace}
\usepackage{enumerate}
\usepackage{amsthm, amsmath, mathtools, amsfonts, amssymb, dsfont, enumitem, mathabx}
\usepackage{bm}
\usepackage[mathscr]{euscript}
\usepackage{adjustbox}
\usepackage{footmisc}

\usepackage[capitalize,noabbrev]{cleveref}
\usepackage{xspace}

\numberwithin{figure}{section}
\numberwithin{table}{section}

\newcommand{\ouracronym}{ODT\xspace}
\newcommand{\Tau}{\mathcal{T}}
\newcommand{\offlinedata}{\mathscr{T}_\text{offline}}
\newcommand{\buffer}{\mathscr{T}_{\text{replay}}}

\newcommand{\set}[1]{\left\{#1\right\}}

\newcommand{\vs}{\mathbf{s}}
\newcommand{\vrtg}{\mathbf{g}}
\newcommand{\va}{\mathbf{a}}

\newcommand{\N}{\mathcal{N}}

\renewcommand{\P}{\operatorname{\mathbb{P}}}
\newcommand{\E}{\operatorname{\mathbb{E}}}

\newcommand{\tb}[1]{\textcolor{blue}{$\bm{#1}$}}
\newcommand{\tp}[1]{\textcolor{Plum}{$\bm{#1}$}}

\newcommand{\hopper}{\texttt{Hopper}\xspace}
\newcommand{\walker}{\texttt{Walker}\xspace}
\newcommand{\cheetah}{\texttt{HalfCheetah}\xspace}
\newcommand{\gym}{\texttt{Gym}\xspace}
\newcommand{\ant}{\texttt{Ant}\xspace}
\newcommand{\antmaze}{\texttt{AntMaze}\xspace}
\newcommand{\medium}{\texttt{medium}\xspace}
\newcommand{\medreplay}{\texttt{medium-replay}\xspace}
\newcommand{\tnll}{T_{\text{NLL}}\xspace}
\newcommand{\tce}{T_{\text{CE}}\xspace}

\graphicspath{{./figure/}}

\theoremstyle{plain}

\theoremstyle{definition}

\theoremstyle{remark}

\usepackage[textsize=tiny]{todonotes}

\usepackage{titlesec}
\icmltitlerunning{Online Decision Transformer}
\begin{document}

\setlength{\textfloatsep}{3pt}
\setlength{\floatsep}{3pt}
\abovedisplayshortskip=0pt
\belowdisplayshortskip=0pt

\titlespacing{\section}{0pt}{0pt}{0pt}
% \titlespacing{\subsection}{0pt}{0pt}{-\parskip}
\titlespacing{\subsection}{0pt}{0pt}{-\parskip}
\titlespacing{\subsubsection}{0pt}{0pt}{-\parskip}
\titlespacing{\paragraph}{0pt}{0pt}{3pt}

\twocolumn[
\icmltitle{Online Decision Transformer}



\icmlsetsymbol{equal}{*}

\begin{icmlauthorlist}
\icmlauthor{Qinqing Zheng}{fair}
\icmlauthor{Amy Zhang}{fair,berkeley}
\icmlauthor{Aditya Grover}{fair,ucla}
\end{icmlauthorlist}

\icmlaffiliation{fair}{Meta AI Research}
\icmlaffiliation{berkeley}{University of California, Berkeley}
\icmlaffiliation{ucla}{University of California, Los Angeles}

\icmlcorrespondingauthor{Qinqing Zheng}{zhengqinqing@gmail.com}

\icmlkeywords{Machine Learning, ICML}

\vskip 0.3in
]



\printAffiliationsAndNotice{}  % leave blank if no need to mention equal contribution

\begin{abstract}
Recent work has shown that offline reinforcement learning (RL) can be formulated as a sequence modeling problem~\cite{chen2021decision,janner2021offline} and solved via approaches similar to large-scale language modeling. However, any practical instantiation of RL also involves an online component, where policies pretrained on passive offline datasets are finetuned via task-specific interactions with the environment. We propose Online Decision Transformers (ODT), an RL algorithm based on sequence modeling that blends offline pretraining with online finetuning in a unified framework. Our framework uses sequence-level entropy regularizers in conjunction with autoregressive modeling objectives for sample-efficient exploration and finetuning. Empirically, we show that \ouracronym is competitive with the state-of-the-art in absolute performance on the D4RL benchmark but shows much more significant gains during the finetuning procedure.
\end{abstract}

\section{Introduction}
Generative pretraining for sequence modeling has emerged as a unifying paradigm for machine learning in a number of domains and modalities, notably in language and vision~\cite{radford2018improving,chen2020generative,brown2020language,lu2021pretrained}.
Recently, such a pretraining paradigm has been extended to offline reinforcement learning (RL)~\cite{chen2021decision,janner2021offline}, wherein an agent is trained to autoregressively maximize the likelihood of trajectories in the offline dataset.
During training, this paradigm essentially converts offline RL to a supervised learning problem~\cite{schmidhuber2019reinforcement,srivastava2019training,emmons2021rvs}.
However, these works present an incomplete picture as policies learned via offline RL are limited by the quality of the training dataset and need to be finetuned to the task of interest via online interactions. It remains an open question whether such supervised learning paradigm can be extended to online settings.

Unlike language and perception, online finetuning for RL is fundamentally different from the pretraining phase as it involves data acquisition via \textit{exploration}.
The need for exploration renders traditional supervised learning objectives (e.g., mean squared error) for offline RL insufficient in the online setting.
Moreover, it has been observed that for standard online algorithms, access to offline data can often have zero or even negative effect on the online performance~\cite{nair2020awac}.
Hence, the overall pipeline for offline pretraining followed by online finetuning for RL policies needs a careful consideration of training objectives and protocols.

We introduce Online Decision Transformers (\ouracronym), a learning framework for RL that blends offline pretraining with online finetuning for sample-efficient policy optimization.
Our framework builds on the decision transformer (DT)~\cite{chen2021decision} architecture previously introduced for offline RL and is especially catered to scenarios where online interactions can be expensive which necessitates both offline pretraining and sample-efficient finetuning.
We identify several key shortcomings with DTs that are incompatible with online learning and rectify them, leading to superior performance for our overall pipeline.

First, we shift from deterministic to stochastic policies for defining exploration objectives during the online phase.
We quantify exploration via the entropy of the policy similar to max-ent RL frameworks~\cite{levine2018reinforcement}.
Unlike traditional frameworks however, the policy entropy for \ouracronym is constrained at an aggregate level over trajectories (as opposed to individual time-steps) and its dual form regularizes a supervised learning objective (as opposed to direct return maximization).
Next, we develop a novel replay buffer~\cite{mnih2015human} consistent with the architecture and training protocol of \ouracronym.
The buffer stores trajectories and is populated via online rollouts from \ouracronym.
Since \ouracronym parameterizes return-conditioned policies, we further investigate strategies for specifying the desired returns during online rollouts.
This value however might not match with the true returns observed during a rollout. To address this challenge, we extend a notion of hindsight experience replay~\cite{andrychowicz_hindsight_2017} to our setting and relabel rolled out trajectories with the corrected return tokens before augmenting them.

Empirically, we validate our overall framework by comparing its performance with state-of-the-art algorithms on the D4RL benchmark~\cite{fu2020d4rl}.
We find that our relative improvements due to our finetuning strategy outperform other baselines~\cite{nair2020awac,kostrikov2021offline}, while exhibiting competitive absolute performance when accounting for the pretraining results of the base model.
Finally, we supplement our main results with rigorous ablations and additional experimental designs to justify and validate the key components of our approach.

\section{Related Work}
Our work encompasses two broad avenues of research which we detail here.
\paragraph{Transformers for RL.} There has been much recent exciting progress formulating the offline RL problem as a context-conditioned sequence modeling problem~\citep{chen2021decision,janner2021offline}. These works build on the reinforcement learning as supervised learning paradigm~\cite{schmidhuber2019reinforcement,srivastava2019training,emmons2021rvs} that focuses on predictive modeling of action sequences conditioned on a task specification (e.g., target goal or returns) as opposed to explicitly learning Q-functions or policy gradients.
% use reward-to-go as context to extract high-reward behavior from a transformer as a generative model. 
\citet{chen2021decision} trains a transformer~\cite{vaswani2017attention,radford2018improving} as a model-free context-conditioned policy, and \citet{janner2021offline} trains a transformer as both a policy and model and show that beam search can be used to improve upon purely model-free performance.
However, these works only explore the offline RL setting, which is similar to the fixed datasets that transformers are traditionally trained with in natural language processing applications. Our work focuses on extending these results to the online finetuning setting, showing competitiveness with state-of-the-art RL methods.

Offline RL methods primarily add a conservative component to an existing off-policy RL method to prevent out-of-distribution extrapolation, but require many tweaks and re-tuning of hyperparameters to work~\citep{NEURIPS2020_0d2b2061,pmlr-v139-kostrikov21a}. Similar to our work, \citet{fujimoto2021minimalist} show the benefits of adding a behavior cloning term to offline RL methods, and that the simple addition of this term allows the porting of off-policy RL algorithms to the offline setting with minimal changes.

\paragraph{Offline RL with Online Finetuning.} While \ouracronym stems from a different perspective than traditional RL methods, there is much existing work focused on the same paradigm of pre-training on a given offline dataset and finetuning in an online environment. 
\citet{nair2020awac} showed that na\"ive application of offline or off-policy RL methods to the offline pre-training and online finetuning regime often does not help, or even hinders, performance. This poor performance in off-policy methods can be attributed to off-policy bootstrapping error accumulation \cite{munos2003error, munos2005error,farahmand2010error, kumar2019stabilizing}. In offline RL methods, poor performance in the online finetuning regime can be explained by excess conservatism, which is necessary in the offline regime to prevent value overestimation of out-of-distribution states. \citet{nair2020awac} was the first to propose an algorithm that works well for both the offline and online training regimes. 
Recent work \cite{kostrikov2021offline} also proposes an expectile-based implicit Q-learning algorithm for offline RL that also shows strong online finetuning performance, because the policy is extracted via a behavior cloning step that avoids out-of-distribution actions. 

\citet{lee2021offlinetoonline} tackles the offline-online setting with a balanced replay scheme and an ensemble of $Q$ functions to maintain conservatism during offline training.   
\citet{lu2021awopt} improves upon AWAC~\citep{nair2020awac}, which exhibits collapse during the online finetuning stage, by incorporating positive sampling and exploration during the online stage.
We also find that positive sampling and exploration are key to good online finetuning, but we will show how these traits naturally occur in \ouracronym, leading to a simple, end-to-end method that automatically adapts to both offline and online settings.
\section{Preliminaries}
\label{sec:prelim}
We assume our environment can be modeled as a Markov decision process (MDP), which can be described as  $\mathcal{M}=\langle\mathcal{S}, \mathcal{A}, p, P, R, \gamma\rangle$, where $\mathcal{S}$ is the state space, $\mathcal{A}$ is the action space, $P(s_{t+1}|s_t, a_t)$ is the probability distribution over transitions, $R(s_t, a_t)$ is the reward function, and $\gamma$ is the discount factor~\citep{bellman1957mdp}.
An agent starts in a initial state $s_1$ sampled from a fixed distribution $p(s_1)$, then at each timestep $t$ it takes an action $a_t \in \mathcal{A}$ from a state $s_t \in \mathcal{S}$ and moves to a next state $s_{t+1} \sim P(\cdot| s_t, a_t)$. After each action the agent receives a deterministic reward $r_t = R(s_{t}, a_t)$.  
Note that our algorithms also directly apply to partially observable Markov decision processes (POMDP), but we use the MDP framework for ease of exposition.

\subsection{Setup and Notation}
\label{sec:notation}
We are interested in online finetuning of Decision Transformer (DT) \cite{chen2020generative}, wherein an agent will have access to a non-stationary training data distribution $\Tau$.
Initially,  during pretraining, $\Tau$ corresponds to the offline data distribution $\Tau_\text{offline}$ and is accessed via an offline dataset $\offlinedata$. During finetuning it is accessed via a replay buffer $\buffer$.
Let $\tau$ denote a trajectory and let $|\tau|$ denote its length. The return-to-go (RTG) of a trajectory $\tau$ at timestep $t$, $g_t = \sum_{t'=t}^{|\tau|} r_{t'}$, is the sum of future rewards from that timestep.
Let $\va = (a_1, \ldots, a_{|\tau|})$, $\vs = (s_1, \ldots, s_{|\tau|})$ and $\vrtg = (g_1, \ldots, g_{|\tau|})$
denote the sequence of action, states and RTGs of $\tau$, respectively.

\subsection{Decision Transformer}
Decision Transformer processes a trajectory $\tau$ as a sequence of 3 types of input tokens: RTGs, states and actions:
$(g_1, s_1, a_1, g_2, s_2, a_2, \ldots, g_{|\tau|}, s_{|\tau|}, a_{|\tau|})$. Specifically, the initial RTG $g_1$ is equal to the return of the trajectory. 
At timestep $t$, DT uses the tokens from the latest $K$ timesteps to generate an action $a_t$. Here, $K$ is a hyperparameter and is also referred to as the \emph{context length} for the transformer. 
Note that the context length during evaluation can be shorter than the one used for training, as we will demonstrate later in our experiments.
DT learns a deterministic policy $\pi_\texttt{DT} (a_t | \vs_{-K, t}, \vrtg_{-K, t})$, where $\vs_{-K, t}$ is shorthand for the sequence of $K$ past states $\vs_{\max(1,t-K+1):t}$ and similarly for $\vrtg_{-K, t}$.
% \footnote{For a timestep $t$ between $1$ and $K$, $\vs_{-K, t} = (s_1, s_2, …, s_t)$ contains $t$ states. For a timestep $t \geq K$, $\vs_{-K, t} = (s_{t-K+1}, s_{t-K+2}, \ldots, s_t)$ contains $K$ states. }. 
This is an autoregressive model of order $K$. In particular, DT parameterized the policy through a GPT architecture~\cite{radford2018improving}, which applies a causal mask to enforce such autoregressive structure in the predicted action sequence.

For simplicity, we assume the data distribution $\Tau$ generates associated length-$K$ action, state and RTG subsequences, which are from the same trajectory. With a little abuse of notation, we also use $(\va, \vs, \vrtg)$ to denote the sample from $\Tau$. This allows us to easily present the training objective of our approach, and the above notations readily apply here. 
The policy is trained to predict the action tokens under the standard $\ell_2$ loss
\begin{equation}
     \E_{(\va, \vs, \vrtg) \sim \Tau} \big[ \tfrac{1}{K} \textstyle \sum_{k=1}^K \displaystyle \big(a_k - \pi_\texttt{DT}(\vs_{-K, k}, \vrtg_{-K, k}) \big)^2 \big].
    \label{eq:obj_dt}
\end{equation}
In practice, we uniformly sample the length-$K$ subsequences from the offline dataset $\offlinedata$ (or the replay buffer $\buffer$ during finetuning), see Appendix~\ref{sec:appendix_sampling}.

For evaluation, we specify the desired performance $g_1$ and an initial state $s_1$. DT then generates the action $a_1 = \pi_\texttt{DT}(s_1, g_1)$. Once an action $a_t$ is generated, we execute it and observe the next state $s_{t+1}\sim P( \cdot \vert s_t, a_t)$ and obtain a reward $r_{t} = R(s_t, a_t)$. This gives us the next RTG as $g_{t+1} = g_t - r_t$. As before, DT generates $a_2$ based on $s_1, s_2$ and $g_1, g_2$. This process is repeated until the episode terminates.

\section{Online Decision Transformer}
\label{sec:odt}
RL policies trained on purely offline datasets are typically sub-optimal due to the limitations of training data as the offline trajectories might not have high return and cover only a limited part of the state space.
One natural strategy to improve performance is to finetune the pretrained RL agents via online interactions.
However, the learning formulation for a standard decision transformer is insufficient for online learning, and as we shall show in our experiment ablations, collapses when used na\"ively for online data acquisition.
% Nonetheless, there are several challenges upfront, including encouraging exploration, generating trajectories with high returns to reuse later, etc.
In this section, we introduce key modifications to decision transformers for enabling sample-efficient online finetuning.

As a first step, we present a generalized, probabilistic learning objective.
We will extend this formulation to account for exploration in online decision transformers (ODT).
In the probabilistic setup, our goal is to learn a \textit{stochastic} policy that maximizes the likelihood of the dataset.
For example, for continuous action spaces, we can use the standard choice ~\cite{sac1, fujimoto2021minimalist, kumar2020conservative, emmons2021rvs}
of a multivariate Gaussian distribution with a diagonal covariance matrix to model the action distributions conditioned on states and RTGs. Let $\theta$ denote the policy parameters. Formally, our policy is
% \\
% \begin{minipage}{\columnwidth}
% \vskip-6pt
% \begin{equation}
% \begin{aligned}
% & && \pi_\theta(a_t | \vs_{-K, t}, \vrtg_{-K,t}) \\
% & = && \N(\mu_\theta(\vs_{-K, t}, \vrtg_{-K,t}), \Sigma_\theta(\vs_{-K, t}, \vrtg_{-K,t})), \; \forall t,
% \end{aligned}
% \end{equation}
% \end{minipage}
%%%% For Arxiv
\begin{equation}
\begin{aligned}
& && \pi_\theta(a_t | \vs_{-K, t}, \vrtg_{-K,t}) \\
& = && \N(\mu_\theta(\vs_{-K, t}, \vrtg_{-K,t}), \Sigma_\theta(\vs_{-K, t}, \vrtg_{-K,t})), \; \forall t,
\end{aligned}
\end{equation}
%%%%
where the covariance matrix $\Sigma_\theta$ is assumed to be diagonal.
Given a stochastic policy, we maximize the log-likelihood of the trajectories in the training dataset, or equivalently minimize the negative log-likelihood (NLL) loss\footnote{\label{note:obj_scale}We scale both $J(\theta)$ and $H^{\Tau}_{\theta}[\va | \vs, \vrtg]$ by $\frac{1}{K}$. As we discuss later, this allows us to easily compare our objective with SAC.}
% \\
% \begin{minipage}{\columnwidth}
% \vskip-8pt
% \begin{equation}
%     \begin{aligned}
%            & J(\theta) = \tfrac{1}{K} \E_{ (\va, \vs, \vrtg) \sim \Tau } [ - \log \pi_\theta(\va | \vs, \vrtg) ] \\
%     % & = && \frac{1}{K}\E_{ (\va, \vs, \vrtg) \sim \Tau }  \bigg(
%     %             - \log \prod_{t=1}^K \pi_\theta(a_t | s_{\leq t}, g_{\leq t})
%     %         \bigg) \\
%      = &\,  \tfrac{1}{K} \E_{ (\va, \vs, \vrtg) \sim \Tau } [ \textstyle -\sum_{k=1}^K \displaystyle \log \pi_\theta(a_k | \vs_{-K, k}, \vrtg_{-K, k})].
%     \end{aligned}
%     \label{eq:sdt_nll}
% \end{equation}
% \end{minipage}
%%%% For Arxiv
\begin{equation}
    \begin{aligned}
           & J(\theta) = \tfrac{1}{K} \E_{ (\va, \vs, \vrtg) \sim \Tau } [ - \log \pi_\theta(\va | \vs, \vrtg) ] \\
    % & = && \frac{1}{K}\E_{ (\va, \vs, \vrtg) \sim \Tau }  \bigg(
    %             - \log \prod_{t=1}^K \pi_\theta(a_t | s_{\leq t}, g_{\leq t})
    %         \bigg) \\
     = &\,  \tfrac{1}{K} \E_{ (\va, \vs, \vrtg) \sim \Tau } [ \textstyle -\sum_{k=1}^K \displaystyle \log \pi_\theta(a_k | \vs_{-K, k}, \vrtg_{-K, k})].
    \end{aligned}
    \label{eq:sdt_nll}
\end{equation}
%%%%
The policies we consider here subsume the deterministic policies considered by DT. Optimizing the objective~\eqref{eq:obj_dt} is equivalent to optimizing~\eqref{eq:sdt_nll} assuming the covariance matrix $\Sigma_\theta$ is diagonal and the variances are the same across all the dimensions, % In other words, $\Sigma_\theta(\vs_{-K, t}, \vrtg_{-K, t}) = \lambda_t I$ for certain positive valued $\lambda_t$, 
which is a special case covered by our assumption.

\subsection{Max-Entropy Sequence Modeling}
\label{sec:odt_model}
The key property of an online RL algorithm is the ability to balance the exploration-exploitation trade-off. Even with stochastic policies, traditional formulation of a DT as in Eq.~\eqref{eq:sdt_nll} does not account for exploration.
To address this shortcoming, we first quantify exploration via the policy entropy\footref{note:obj_scale} defined as:
\begin{equation}
\begin{aligned}
    & H^{\Tau}_{\theta}[\va | \vs, \vrtg]  = \tfrac{1}{K} \E_{(\vs, \vrtg) \sim \Tau} \big[  H[\pi_\theta(\va | \vs, \vrtg)]  \big] \\
    =  & \tfrac{1}{K} \E_{(\vs, \vrtg) \sim \Tau}\big[ \textstyle \sum_{k=1}^K H[\pi_\theta(a_k | \vs_{-K, k}, \vrtg_{-K, k}) ]\big],
    % & = && \frac{1}{K}  \E_{(\vs, \vrtg) \sim \Tau} \bigg(
    %              \int_{\vx \sim \pi_\theta} -\log \pi_\theta(x | \vs, \vrtg )
    %         \bigg) \\
    % & = && \frac{1}{K}  \E_{(\vs, \vrtg) \sim \Tau}\bigg(
    %              \int_{\vx \sim \pi_\theta} -\log \prod_{t=1}^K  \pi_\theta(x _t | s_{\leq t}, g_{\leq t})
    %         \bigg) \\
    % =  & \frac{1}{K}\E_{(\vs, \vrtg) \sim \Tau}\big[ \int_{\vx \sim \pi_\theta} -\sum_{t=1}^K \log \pi_\theta(x_t | \vs_{-K, t}, \vrtg_{-K, t}) \big].
    \end{aligned}
    \label{eq:sdt_entropy}
\end{equation}
where $H[\pi_\theta(a_k)]$ denotes the Shannon entropy of the distribution $\pi_\theta(a_k)$. The policy entropy depends on the data distribution $\Tau$, which is static in the offline pretraining phase but dynamic during finetuning as it depends on the online data acquired during exploration.

Similar to many existing max-ent RL algorithms~\cite{levine2018reinforcement} such as Soft Actor Critic (SAC, \citet{sac1, sac2}),
we explicitly impose a lower bound on the policy entropy to encourage exploration.
That is, we are interested in solving the following constrained problem:
% \\
% \begin{minipage}{\columnwidth}
% \begin{equation}
%     \min_\theta J(\theta) \;\; \text{subject to} \;\; H^{\Tau}_{\theta}[\va | \vs, \vrtg] \geq \beta,
%     \label{eq:sdt_main}
% \end{equation}
% \end{minipage}
%%% For Arxiv
\begin{equation}
    \min_\theta J(\theta) \;\; \text{subject to} \;\; H^{\Tau}_{\theta}[\va | \vs, \vrtg] \geq \beta,
    \label{eq:sdt_main}
\end{equation}
%%%
where  $\beta$ is a prefixed hyperparameter.
Following \citet{sac2}, in practice, we solve the dual problem of \eqref{eq:sdt_main} to avoid explicitly dealing with the inequality constraint.
Namely, we consider the Lagrangian $L(\theta, \lambda) = J(\theta) + \lambda (\beta - H^{\Tau}_{\theta} [\va | \vs, \vrtg])$ and solve the problem $\displaystyle \max_{\lambda \geq 0} \min_{\theta} L(\theta, \lambda)$ by
alternately optimizing $\theta$ and $\lambda$. Optimizing $\theta$ with fixed $\lambda$ is equivalent to
% \\
% \begin{minipage}{\columnwidth}
% \begin{equation}
%     \min_{\theta}\, J(\theta) - \lambda H^{\Tau}_{\theta} [\va | \vs, \vrtg],
%     \label{eq:sdt_opt_theta}
% \end{equation}
% \end{minipage}
% and optimizing $\lambda$ with fixed $\theta$ boils down to solving\\
% \begin{minipage}{\columnwidth}
% \begin{equation}
%     \min_{\lambda \geq 0} \, \lambda (H^{\Tau}_{\theta}[\va|\vs, \vrtg] - \beta).
%     \label{eq:sdt_opt_lambda}
% \end{equation}
% \end{minipage}
%%% For Arxiv
\begin{equation}
    \min_{\theta}\, J(\theta) - \lambda H^{\Tau}_{\theta} [\va | \vs, \vrtg],
    \label{eq:sdt_opt_theta}
\end{equation}
and optimizing $\lambda$ with fixed $\theta$ boils down to solving
\begin{equation}
    \min_{\lambda \geq 0} \, \lambda (H^{\Tau}_{\theta}[\va|\vs, \vrtg] - \beta).
    \label{eq:sdt_opt_lambda}
\end{equation}
%%%
We highlight a few focal points for understanding our method from both the primal and dual point of view.
% \vspace*{-1em}
% \begin{enumerate}[leftmargin=*]\itemsep0em
%     \item

  (1) Problem~\eqref{eq:sdt_opt_theta} can be viewed as the corresponding regularized form of our primal problem~\eqref{eq:sdt_main}, where the dual variable $\lambda$ serves the role of temperature variable in many regularized RL formulations.
    As a key difference to SAC and other classic RL methods, our loss function $J(\theta)$ is the NLL rather than the discounted return. In principal, we focus only on supervised learning of action sequences as opposed to explicitly maximizing returns.
    % our method is an entirely supervised learning algorithm without targeting return maximization.
    Consequently, the objective in Eq.~\eqref{eq:sdt_opt_theta}
    % $J(\theta) - \lambda H^{\Tau}_{\theta}[\va | \vs, \vrtg]$
    can be interpreted as minimizing the expected difference between the log-probability of the observed actions and $\lambda$-scaled log-probability of actions drawn from $\pi_\theta(\cdot|\vs, \vrtg)$. That is, we attempt to learn $\pi_\theta$ so that it matches the observed action distribution with some deviation, and the dual variable $\lambda$ explicitly controls the degree of mismatch.

    % \item
    (2) Rigorously speaking, $H^{\Tau}_{\theta}[\va | \vs, \vrtg]$ is a cross conditional entropy. This is because the training data distribution $\Tau$ is generally not the same as the action-state-RTG marginals induced by the current policy $\pi_\theta$ and the transition probability $P$. During pretraining, $\Tau$ is the fixed offline data distribution $\Tau_\text{offline}$ whereas during finetuning, as we use a replay buffer to store the collected online data, $\Tau$ is accessed via a replay buffer that depends on the current policy $\pi_\theta$ and the data gathered by the policies at the previous iterations.
    However, over the course of training, if the policy converges, this cross entropy term will converge to the entropy.
    Section~\ref{sec:odt_convergence} discusses this in more detail. In other words, our objective function automatically adapts to play a suitable role for both offline and online settings. During offline training, the cross entropy term controls the degree of distribution mismatch whereas during online training, it becomes the entropy and encourages policy exploration.

    % \item
  (3)  Another important difference to the classic max-ent RL methods is that our policy entropy $H^{\Tau}_{\theta}[\va|\vs, \vrtg]$, as in Eq.~\eqref{eq:sdt_entropy}, is defined at the level of sequences, as opposed to transitions.
  Consequently, our constraint in the primal problem~\eqref{eq:sdt_main} differs from the entropy constraint for SAC~\cite{sac2}. For simplicity, let us ignore the RTG variable $\vrtg$ in our policy entropy. While SAC imposes a lower bound $\beta$ for the policy entropy at all timesteps, our entropy $H^{\Tau}_{\theta}[\va|\vs, \vrtg]$ is averaged over $K$ consecutive timesteps.
    Hence, our constraint only requires that the entropy averaged over a sequence of $K$ timesteps is lower bounded by $\beta$.
    Therefore, any policy satisfying the transition-level policy constraint of SAC also satisfies our sequence-level constraint. Hence, the space of feasible policies is larger in our case whenever $K>1$. When $K=1$, the sequence-level constraint reduces to the transition-level constraint of SAC.

% \end{enumerate}

Finally, we make several remarks regarding similarities to SAC with respect to the practical optimization details. First, we do not fully solve the sub-problems \eqref{eq:sdt_opt_theta} and \eqref{eq:sdt_opt_lambda}. For both of them, we only take one gradient update each time, a.k.a one-step alternating gradient descent. Second, the evaluation of $H^{\Tau}_{\theta}[\va|\vs, \vrtg]$ involves integrals.
% \footnote{ $H[\pi_\theta(a_t |\vs_{-K, t}, \vrtg_{-K, t})] = \int_{\vx \sim \pi_\theta} - \log \pi_\theta(x_t | \vs_{-K, t}, \vrtg_{-K, t})$.}.
We approximate each integral using the one-sample Monte Carlo estimation,
and the sample is re-parameterized for low-variance gradient computation. As also noted by \citet{sac2}, we often observe that the constraint in problem~\eqref{eq:sdt_main} is tight so that the actual entropy $H^{\Tau}_{\theta}[\va | \vs, \vrtg]$ matches $\beta$.

\subsection{Training Pipeline}
\label{sec:odt_pipeline}

We instantiate the formulation described above using a transformer architecture. The online decision transformer (ODT) builds on the DT architecture and incorporates changes due to the stochastic policy. We predict the policy mean and log-variance
% $\mu_\theta$ and $\log \text{diag}(\Sigma_\theta)$
by two separate fully connected layers at the output.
% on top of the Transformer output.
Algorithm~\ref{alg:odt} summarizes the overall finetuning pipeline in \ouracronym, where the detailed inner training steps are described in Algorithm~\ref{alg:dt-train}.
% Our procedure is simple, proximal to but different from tradition methods in a few aspects.
We outline the major components of these algorithms below and discuss additional design choices and hyperparameters in Appendix~\ref{sec:appendix_design}.

\paragraph{Trajectory-Level Replay Buffer.} We use a replay buffer to record our past experiences and update it periodically.
For most existing RL algorithms, the replay buffer is composed of transitions.
After every step of online interaction within a rollout,
the policy or the Q-function is updated via a gradient step and the policy is executed to gather a new transition for addition
into the replay buffer. For \ouracronym however, our replay buffer consists of trajectories rather than transitions. After offline pretraining, we initialize the replay buffer by the trajectories with highest returns in the offline dataset. Every time we interact with the environment, we completely rollout an episode with the current policy, then refresh the replay buffer using the collected trajectory in a first-in-first-out manner. Afterwards, we update the policy and rollout again, as shown in Algorithm~\ref{alg:odt}.
Similar to \citet{sac1}, we also observe that evaluating the policy using the mean action generally leads to a higher return, but it is beneficial to use sampled actions for online exploration since it generates more diverse data.
% Similar observations has been reported in \citet{sac1}.
\begin{algorithm}[t]
 \DontPrintSemicolon
 \small
  \textbf{Input:} offline data $\offlinedata$, rounds $R$,
   exploration RTG $g_\text{online}$, buffer size $N$, gradient iterations $I$, pretrained policy $\pi_\theta$ \;
   \textbf{Intialization:} Replay buffer $\buffer \leftarrow$ top $N$ trajectories in $\offlinedata$.\; \label{alg:odt_buffer_init}
  \For{round = $1, \ldots, R$}{
    \tcp{use randomly sampled actions}
    Trajectory $\tau \leftarrow$ Rollout using $\mathcal{M}$ and $\pi_\theta(\cdot | \vs, \vrtg(g_\text{online}))$.\; \label{alg:odt_gonline}
    $\buffer\leftarrow \left\{\buffer \backslash \set{\text{the oldest trajectory}} \right\} \bigcup \set{\tau}$.\;
    $\pi_\theta \leftarrow$ Finetune ODT on $\buffer$ for $I$ iterations via Algorithm~\ref{alg:dt-train}.\;
  }
  \caption{Online Decision Transformer}
\label{alg:odt}
\end{algorithm}

\paragraph{Hindsight Return Relabeling.} Hindsight experience replay (HER) is a method for improving the sample-efficiency of goal-conditioned agents in environments with sparse rewards~\cite{andrychowicz_hindsight_2017, rauber2017hindsight, ghosh2019learning}.
The key idea here is to relabel the agent's trajectories with the achieved goal, as opposed to the intended goal.
% The aim of those problems is usually to develop a goal-conditioned policy so that the agent is capable of reaching various goals. Under this setting, relabeling
% the goals and associated rewards of trajectories in hindsight can greatly improve data efficiency.
In the case of \ouracronym, we are learning policies conditioned on an initial RTG.
The return achieved during a policy rollout and the induced RTG can however differ from the intended RTG.
Inspired by HER, we relabel the RTG token for the rolled out trajectory $\tau$ with the achieved returns, such that the RTG token at the last timestep $g_{|\tau|}$ is exactly the reward obtained by the agent $r_{|\tau|}$, see Line~\ref{alg:odt_hindsight} of Algorithm~\ref{alg:dt-train}.
This return relabeling strategy applies to environments with both sparse and dense rewards.

\begin{algorithm}[t]
 %\SetAlgoLined
 \small
 \DontPrintSemicolon
  \textbf{Input:} model parameters $\theta$, replay buffer $\buffer$, training iterations $I$, context length $K$, batch size $B$\;
Compute the trajectory sampling probability $p({\tau}) = |\tau| / \sum_{\tau \in \Tau}|\tau|.$

\For{$t = 1, \ldots, I$}{
    Sample $B$ trajectories out of $\buffer$ according to $p$. \;
    \For{each sampled trajectory $\tau$} {
        \tcp{Hindsight Return Relabeling}
        $\vrtg \leftarrow$ the RTG sequence computed by the true rewards: $\vrtg_t = \sum_{j=t}^{|\tau|} r_j, \; 1\leq t \le |\tau|.$\; \label{alg:odt_hindsight}
        $(\va, \vs, \vrtg) \leftarrow$ a length $K$ sub-trajectory uniformly sampled from $\tau$. \;
    }
    $\theta \leftarrow $ one gradient update using the sampled $ \set{(\va, \vs, \vrtg)}$s.
}
  \caption{ODT Training}
\label{alg:dt-train}
% \ag{i think we should enable line numberings inside the algorithms and explictly refer to those in the text so reader knows what line is being discussed in which paragraph}
\end{algorithm}

\paragraph{RTG Conditioning.}
\ouracronym requires an hyperparameter, initial RTG $g_\text{online}$, for gathering additional online data (see Line~\ref{alg:odt_gonline} of Algorithm~\ref{alg:odt}). Previously,
\citet{chen2021decision} showed that the actual evaluation return of offline DT has strong correlation with the initial RTG empirically and can often extrapolate to RTG values beyond the maximum returns observed in the offline dataset.
% showed that for some environments, DT is even capable of extrapolation when provided unseen RTG values.
For ODT, we find it best to set this hyperparameter to a small, fixed scaling (set to 2 in our experiments) of the expert return.
We also experimented with much larger values as well as time-varying curriculum (e.g., quantiles of the best evaluation return in the offline and online datasets) but we found these to be suboptimal relative to a fixed, scaled RTG.
% version of the
% We have similar observations for ODT, see Section~\ref{sec:expr_ablate}. As a result, for all the tasks we consider throughout this paper, we use an out of distribution initial RTG $g_\text{online}$ in online exploration, which is 2x the expert performance for each particular task.

\paragraph{Sampling Strategy.}
Similar to DT, Algorithm~\ref{alg:dt-train} uses a two step sampling procedure to ensure that the sub-trajectories of length $K$ in the replay buffer $\buffer$ are sampled uniformly. We first sample a single trajectory with probability proportional to its length, then uniformly sample a sub-trajectory of length $K$. For environments with non-negative dense rewards, our sampling strategy is akin to importance sampling. In those cases, the length of a trajectory is highly correlated with its return, as we further highlight in Appendix~\ref{sec:appendix_sampling}.

\subsection{Training Dynamics}
\label{sec:odt_convergence}
\begin{figure}
    \centering
    \includegraphics[width=\columnwidth]{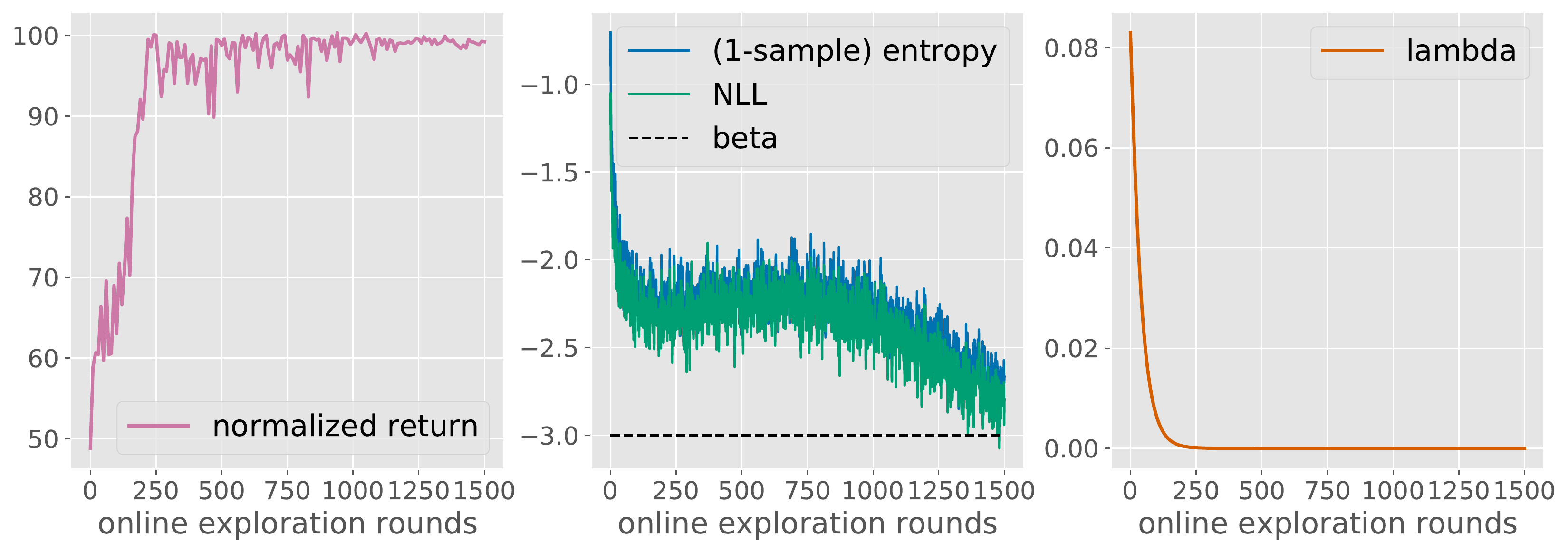}
    % %%% For Arxiv
    % \includegraphics[width=0.8\columnwidth]{./figure/odt_convergence_1row.pdf}
    % %%%
    \vskip-10pt
    \caption{An example run of ODT on \hopper task using \medium dataset. ODT's evaluation return converges over the course of training. Meanwhile, both the NLL and the entropy are converging to $\beta$, which is $-3$ in this run. The dual variable $\lambda$ converges to zero.}
    \label{fig:odt_convergence_example}
\end{figure}

We comment on some empirical observations regarding the training dynamics of \ouracronym and their implications.
We first show an example run where the online return of \ouracronym saturates, suggesting that training has converged.
Restricting ourselves to cases where Algorithm~\ref{alg:odt} converges, we discuss the training dynamics of ODT.
This convergence assumption enables us to analyze the simplification of the learning objective over the course of training,
and also the change in behavior of the initial RTG token for ODT policies.
We emphasize that the convergence guarantee of Algorithm~\ref{alg:odt} is an open question and beyond the scope of this paper, and our claims will be guided primarily via experiments.

Figure~\ref{fig:odt_convergence_example} plots an example run of ODT. The left panel shows that
the return of ODT converges, suggesting a potential convergence of the evaluation policy $\pi_\theta(\cdot|\vs, \vrtg_\text{eval})$,
where $\vrtg_\text{eval}$ is the RTG sequence induced by the initial RTG token $g_\text{eval}$ for evaluation rollouts.

Let us consider the objective function in Eq.~\eqref{eq:sdt_opt_theta}: 
\vspace*{-0.5em}
\begin{equation}
\resizebox{.9\columnwidth}{!}{
$\min_\theta \overbrace{\E_{(\vs, \va, \vrtg) \sim \Tau} \left[ - \log \pi_\theta(\va | \vs, \vrtg) \right]}^{
{\tnll}
}
 - \lambda \overbrace{\E_{(\vs, \vrtg) \sim \Tau} [H(\pi_\theta(\cdot | \vs, \vrtg))]}^{\tce}$.
 }
 \label{eq:sdt_convergence_main}
\end{equation}

% %%% For Arxiv
% \begin{equation}
% \min_\theta \overbrace{\E_{(\vs, \va, \vrtg) \sim \Tau} \left[ - \log \pi_\theta(\va | \vs, \vrtg) \right]}^{
% {\tnll}
% }
%  - \lambda \overbrace{\E_{(\vs, \vrtg) \sim \Tau} [H(\pi_\theta(\cdot | \vs, \vrtg))]}^{\tce}.
%  \label{eq:sdt_convergence_main}
% \end{equation}
% %%%
As discussed previously in Section~\ref{sec:odt_model}, the training data distribution $\Tau$ is static in the offline training stage, but keeps evolving during finetuning. Problem~\eqref{eq:sdt_convergence_main} falls into the class of optimization problems where the data distribution is non-stationary and depends on the model parameters. Recent advances in optimization show that the iterates of those problems converge to a fixed point with (stochastic) gradient updates under certain conditions, e.g., see related works including \citet{bottou2013counterfactual, perdomo2020performative, mendler2020stochastic, drusvyatskiy2020stochastic}.
Motivated by our empirical observations and the recent theoretical progress, in the rest of this section, we analyze the training dynamics of \ouracronym under the assumption that the policy learned via Algorithm~\ref{alg:odt} converges.

To start, we discuss how problem~\eqref{eq:sdt_convergence_main} evolves. Its objective contains two terms, $\tnll$ and $\tce$. As mentioned in Section~\ref{sec:odt_model}, $\tce$ is a cross entropy during pretraining because the expectation is w.r.t. $\Tau = \Tau_\text{offline}$ (the offline data distribution) rather than the marginal action-state-RTG distribution $\rho_{\pi_\theta}(\va, \vs, \vrtg)$ induced by $\pi_\theta$ and $P$.  As \ouracronym training converges, a few consequences follow. For simplicity, let us ignore the hindsight return relabeling for now. First, in the online finetuning stage, the training data is sampled from the replay buffer which contains the past exploration rollouts. As the policy $\pi_\theta$ converges, $\Tau$ will also converge to the policy induced marginal $\rho_{\pi_\theta}(\va, \vs, \vrtg)$. If this happens, the cross entropy term $\tce$ reduces to the conditional entropy
$H_{\rho_{\pi_\theta}}[\va | \vs, \vrtg]$, which is also equal to the NLL term $\tnll$.

As a result, problem~\eqref{eq:sdt_convergence_main} reduces to NLL minimization if $\lambda$ converges to a value between $0$ and $1$ ($0 \leq \lambda < 1$):
% \qq{actually, such reduction will happen if $\lambda < 1$.}:
\begin{equation}
 \min_\theta \, \E_{(\vs, \va, \vrtg) \sim \rho_{\pi_\theta}} \left[ - \log \pi_\theta(\va | \vs, \vrtg) \right].
\label{eq:sdt_convergence_online}
\end{equation}

By the complementary slackness~\cite{boyd2004convex}, the objective of problem~\eqref{eq:sdt_opt_lambda} converges to zero. For the example showed in Figure~\ref{fig:odt_convergence_example}, the constraint is inactive i.e., the entropy is always larger than $\beta$, therefore, the dual variable $\lambda$ converges to zero. We have also observed cases where the constraint is (stochastically) tight, for which $\lambda$ can be positive. In our experiments, we consistently observe that $\lambda$ converges to a value between $0$ and $1$ even when starting from various initial values. In both scenarios, the overall loss will converge to the desired formulation~\eqref{eq:sdt_convergence_online}. See Appendix~\ref{sec:appendix_lambda_convergence} for more discussions.

% and the iterates converge to a fixed point.
Put differently, the above objective performs behavior cloning on trajectories generated as per $\rho_{\pi_\theta}(\va, \vs, \vrtg)$.
% Next, we discuss the convergence of ODT policies.
Now, let us consider hindsight return relabeling and take a closer look at problem~\eqref{eq:sdt_convergence_online}.
% Suppose we condition on an expert-performance valued RTG $g_\text{online}$ for exploration and leverage the hindsight return relabeling technique.
Let $\vrtg_\text{online}$ denote the RTG subsequence induced by the exploration RTG $g_\text{online}$, and $\vrtg_\text{real}$ denote the real RTG subsequence obtained by relabeling: $\vrtg_{\text{real}, t} = \sum_{i=t}^{|\tau|} r_t $. Problem~\eqref{eq:sdt_convergence_online} now becomes
\begin{equation}
 \min_\theta \, \E_{\vs \sim P, \va \sim \pi_\theta( \cdot | \vs, \vrtg_\text{online})} \left[ - \log \pi_\theta(\va | \vs, \vrtg_\text{real}(\vs, \va)) \right].
\label{eq:sdt_convergence_online_hindsight}
\end{equation}
This formulation suggests that we are trying to match the policies $\pi_\theta( \cdot | \vs, \vrtg_\text{real})$ for all the observed $\vrtg_\text{real}$ with a single policy $\pi_\theta( \cdot | \vs, \vrtg_\text{online})$.

\begin{figure}[t]
    \centering
    \includegraphics[height=0.5\columnwidth]{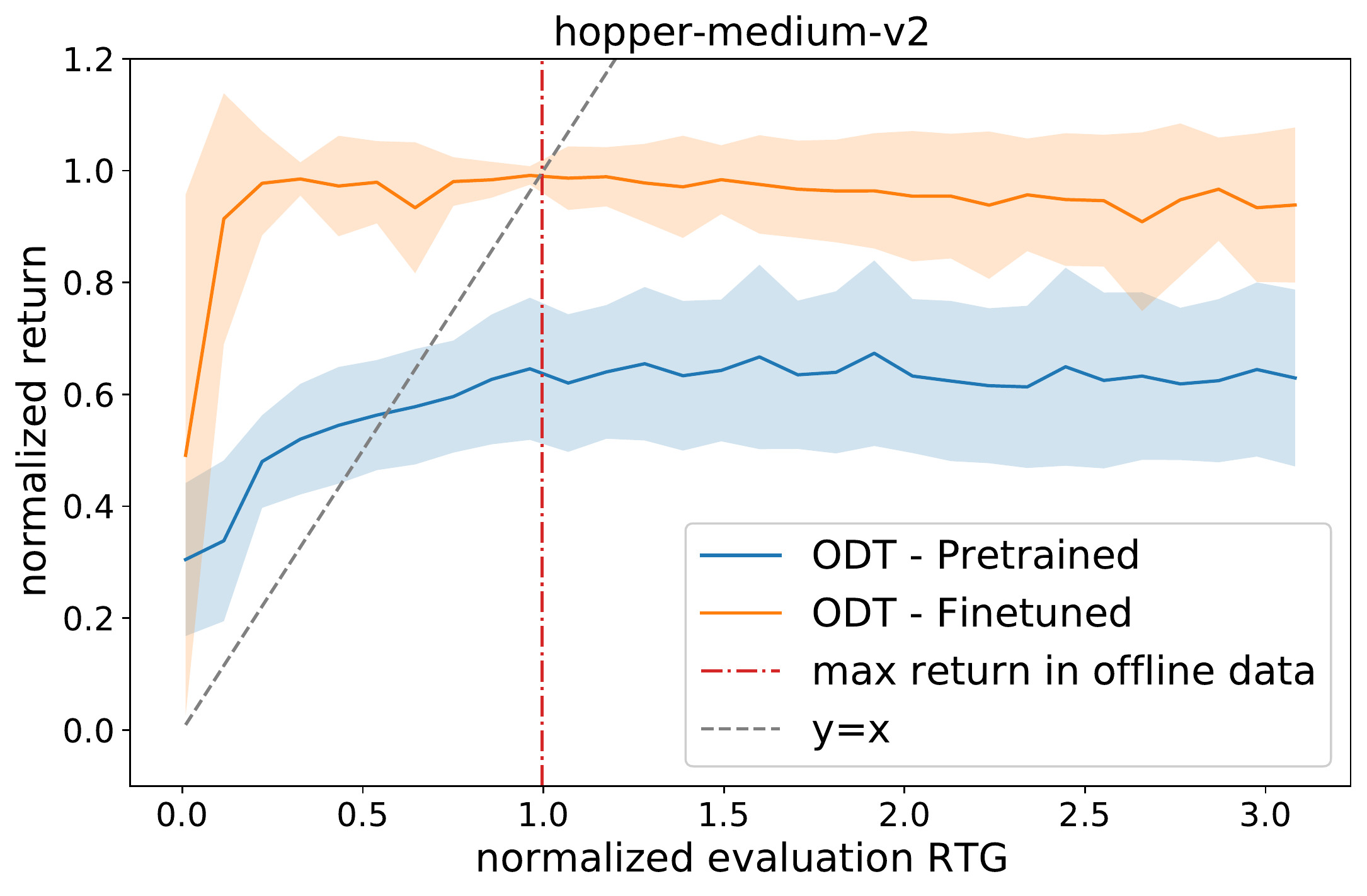}
    %%% For Arxiv
    % \includegraphics[width=0.5\columnwidth]{figure/ablation_rtg_hopper_medium.pdf}
    \caption{The evaluation returns of ODT when conditioned on a wide range of values for $g_\text{eval}$.}
    \label{fig:odt_convergence_rtg}
\end{figure}

 Empirically, we make observations in Figure~\ref{fig:odt_convergence_rtg} that coincide with the above analysis. Here, we vary the initial evaluation RTG $g_\text{eval}$ for \hopper and inspect the return of ODT after pretraining on \medium dataset and finetuning, respectively. The return of ODT is strongly correlated with $g_\text{eval}$ after the offline pretraining, although the performance saturates after a certain threshold of $g_\text{eval}$ (around $0.85$ in Figure~\ref{fig:odt_convergence_rtg}).
 In comparison, the finetuned ODT is less sensitive to $g_\text{eval}$ except for exceptionally large values, where the return slightly declines and the variance increases (cf. Section~\ref{sec:expr_ablate}). Besides, the threshold for performance saturation gets pushed to an extremely low value (around $0.1$ in Figure~\ref{fig:odt_convergence_rtg}).
This demonstrates that while the offline pretrained ODT models a relatively wide distribution of policy returns, the distribution learned by ODT after finetuning concentrates on a narrow range of returns, suggesting a potential concentration on a single policy.

\section{Experiments}
\label{sec:expr_comp}

Our experiments aim to answer two primary questions:
% \vspace*{-1em}
\begin{enumerate}[leftmargin=*]\itemsep0em
    \item How does \ouracronym compare with other state-of-the-art approaches for finetuning pretrained policies under a \emph{limited} online budget?
    \item How do the individual components of \ouracronym influence its overall performance?
\end{enumerate}

\paragraph{Tasks and Datasets.} For answering both these questions, we focus on two types of tasks with offline datasets from the D4RL benchmark \cite{fu2020d4rl}. The first type consists of the \gym locomotion tasks \hopper, \walker, \cheetah, and \ant, which are standard environments with dense rewards. For these environments, our evaluation uses the v2 \medium and \medreplay datasets which contain trajectories collected by sub-optimal policies. The \medium dataset contains 1M samples from a policy trained to approximately $\frac{1}{3}$ the performance of an expert policy, and the \medreplay dataset uses the replay buffer of a policy trained up to the performance of a \medium agent. For the second setup, we consider the goal-reaching tasks \antmaze, where the objective is to move an Ant robot to a target location and the rewards are sparse. The agent obtains reward $1$ if it reaches the goal location otherwise $0$. We use the v2 \texttt{umaze} and \texttt{umaze-diverse} datasets in our experiments.

\subsection{Benchmark Comparisons}
For a thorough understanding of different methods, we compare both the offline and online performance.
We compare the offline performance of \ouracronym with DT and implicit Q-learning (IQL) \cite{kostrikov2021offline}, a state-of-the-art algorithm for offline RL.
We also compare our online finetuning performance to the finetuning variant of IQL, which essentially incorporates the advantage weighted actor critic (AWAC) \cite{nair2020awac} technique for finetuning. For a purely online baseline, we also report the results of soft actor critic (SAC) \cite{sac1} at $200$k online steps. We use the official Pytorch implmentation\footnote{\scriptsize \url{https://github.com/kzl/decision-transformer}} for DT, the official JAX implementation\footnote{\scriptsize \url{https://github.com/ikostrikov/implicit_q_learning}} for IQL, and the Pytorch implementation\footnote{\scriptsize \url{https://github.com/denisyarats/pytorch_sac}} \cite{pytorch_sac} for SAC.

\paragraph{Hyperparameters.} We use the default hyperparameters in the open-source codebase for DT and IQL, and those in \citet{sac1} for SAC.\footnote{The hyperparameters in the Pytorch SAC codebase are optimized for \texttt{dm-control} tasks. We found those used in \citet{sac1} lead to better results.}
 Following the setup of \citet{kostrikov2021offline}, the replay buffer we use for IQL and SAC can contain up to 1 million transitions for \gym tasks and 2 million transitions for \antmaze. To make a fair comparison, we restrict the size of the \ouracronym replay buffer so that the maximum number of transitions matches with IQL and SAC, which is $1000$ for \gym and $2857$ for \antmaze. However, we observe that a smaller replay buffer is beneficial for \antmaze and we use $1500$ in our experiments. IQL and SAC both make one gradient update after each online step, and \ouracronym runs $300$ gradient updates between two consecutive exploration rollouts.
 The complete list of hyperparameters of \ouracronym are summarized in Appendix~\ref{sec:appendix_hp}.

\begin{table*}[t]
    \centering
    %\tiny
    % \scriptsize
    \resizebox{\textwidth}{!}{
    \begin{tabular}{l || l  l  l  l  l  l l c  }
    \toprule
         dataset & DT & \ouracronym (offline) &  \ouracronym (0.2m)  & $\delta_{\text{\ouracronym}}$ &  IQL (offline) &  IQL (0.2m) & $\delta_{\text{IQL}}$ &   SAC (0.2m) \\ \midrule
         %\cmidrule{2-10}
         % & offline   & offline &  0.2m &  1m  &  offline & 0.2m &  1m &  0.2m &  1m \\ \midrule
        hopper-medium & $61.03 \pm 5.11$ & $66.95 \pm 3.26$ & \tb{97.54 \pm 2.10} & \tb{30.59} & $63.81 \pm9.15$ & $66.79 \pm  4.07$ & $2.98$ &  $66.76 \pm 27.13$ \\
        hopper-medium-replay & $62.75 \pm 15.05$ & $86.64\pm5.41$ & $88.89\pm 6.33$ & $2.25$ & $92.13 \pm 10.43$ & \tb{96.23  \pm  4.35} & \tb{4.10} &  \\

        walker2d-medium & $72.03 \pm 4.32$ & $72.19 \pm 6.49$ & $76.79 \pm 2.30$ & \tb{4.60} & $79.89 \pm 3.06$ & \tb{80.33 \pm 2.33} & $0.44$ & $34.20 \pm 18.43$\\
        walker2d-medium-replay & $42.53 \pm 15.36$ & $68.92 \pm 4.79$ & \tb{76.86 \pm 4.04} & \tb{7.94} & $73.67 \pm 6.37$ & $70.55 \pm 5.81$ & $-3.12$ & \\

        halfcheetah-medium & $42.43 \pm 0.30$ & $42.72 \pm 0.46$ & $42.16 \pm 1.48$ & $-0.56$ &$47.37 \pm 0.29$ & $47.41 \pm 0.15$ & \tb{0.04} & \tb{55.73 \pm 4.19}\\
        halfcheetah-medium-replay & $35.92 \pm 1.56$ & $39.99 \pm 0.68$ & $40.42 \pm 1.61$ & \tb{0.43}     &$44.10 \pm 1.14$ & $44.14 \pm 0.3$  & $0.04$\\
        ant-medium & $ 93.56 \pm 4.94$  & $91.33 \pm 4.13$  & $90.79 \pm 5.80$ &  $-0.54$    & $99.92 \pm 5.86$ & \tb{100.85 \pm 2.02} &   \tb{0.93}   & $30.03 \pm 7.98$\\
        ant-medium-replay & $89.08 \pm 5.33$ & $86.56 \pm 3.26$ &  \tb{91.57 \pm 2.73} &  \tb{5.01}    & $91.21 \pm 7.27$ & $91.36 \pm 1.47$ & $0.15$ \\ \hline
        sum & & & \tb{605.02} & \tb{49.72} &  & 597.66 & 5.56\\ \midrule

        antmaze-umaze & $53.3 \pm 5.52$ & $53.10 \pm 4.21$ & \tp{88.5 \pm 5.88} &  \tb{35.4}  & $87.1 \pm 2.81$ & \tp{89.5 \pm 5.43} & $2.4$  & /  \\
        antmaze-umaze-diverse & $52.5 \pm 9.89$ & $50.20 \pm 6.69$ & \tp{56.00 \pm 5.69} &   \tb{7.99}   & $64.4 \pm 8.95$ & \tp{56.8 \pm 6.42} & $-7.6$ & /\\ \hline
        sum & & & \tp{144.5} & \tb{43.39} &  & \tp{146.3} & $-5.2$\\
        \bottomrule
    \end{tabular}
    }
    \caption{Comparison of the average normalized returns on \gym and \antmaze tasks. We report the mean and standard deviation over $10$ seeds.
    Blue: The best performance with $200$k online samples. Purple: \ouracronym achieves nearly the same performance as IQL.
    }
    \label{tbl:finetune}
\end{table*}
\paragraph{Results and Analysis.} For each method, we train $10$ instances with different seeds. For each instance, we run $10$ evaluation trajectories for \gym tasks and $100$ ones for \antmaze tasks, respectively. Table~\ref{tbl:finetune} report the results.

For simply offline pretraining, IQL outperforms both \ouracronym and DT on most tasks and datasets. However, with an additional budget of $200$k online samples, we observe notable performance improvement for \ouracronym on \hopper, \walker, and \antmaze and the relative performance gap $\delta_\text{\ouracronym}$ between the offline and online phase is significant for \ouracronym.
In contrast, IQL only shows limited improvements $\delta_\text{IQL}$ given the same online budget.
On average, the relative improvements due to finetuning for ODT ($\delta_\text{\ouracronym}$) are $\sim$9x to those of IQL ($\delta_\text{IQL}$) for the \gym tasks and datasets.
The average absolute performance is similar for both approaches, even though IQL had a much better pretrained initialization to begin with.
We also obtain significant improvements due to finetuning on the challenging \ant environments and match IQL's absolute performance.
Here, we note that IQL struggles to improve its performance and moreover, its finetuning protocol could also degrade the initial performance (e.g., \antmaze with \texttt{umaze-diverse} dataset).

Finally, it is also instructive to view the results through the lens of online training. We report the performance of SAC, the representative baseline of online RL methods
under a sample budget of $200$k online interactions. SAC performs substantially worse than \ouracronym on all the \gym tasks except \cheetah. In addition, we found that SAC fails to learn non-trivial policies for \antmaze.
While it still remains an open question that if transformer-based model-free RL methods can be learned purely online, our results suggest that \ouracronym can significantly benefit in practical regimes with offline data and limited budget for online interactions.

\subsection{Ablation Study}
\label{sec:expr_ablate}
We ablate the design choices for \ouracronym to identify the components that are key to its performance. Due to the lack of space, we defer additional experiments to Appendix~\ref{sec:appendix_additioanl_ablation}.

\begin{figure}[t]
    \centering
    \includegraphics[width=0.48\columnwidth]{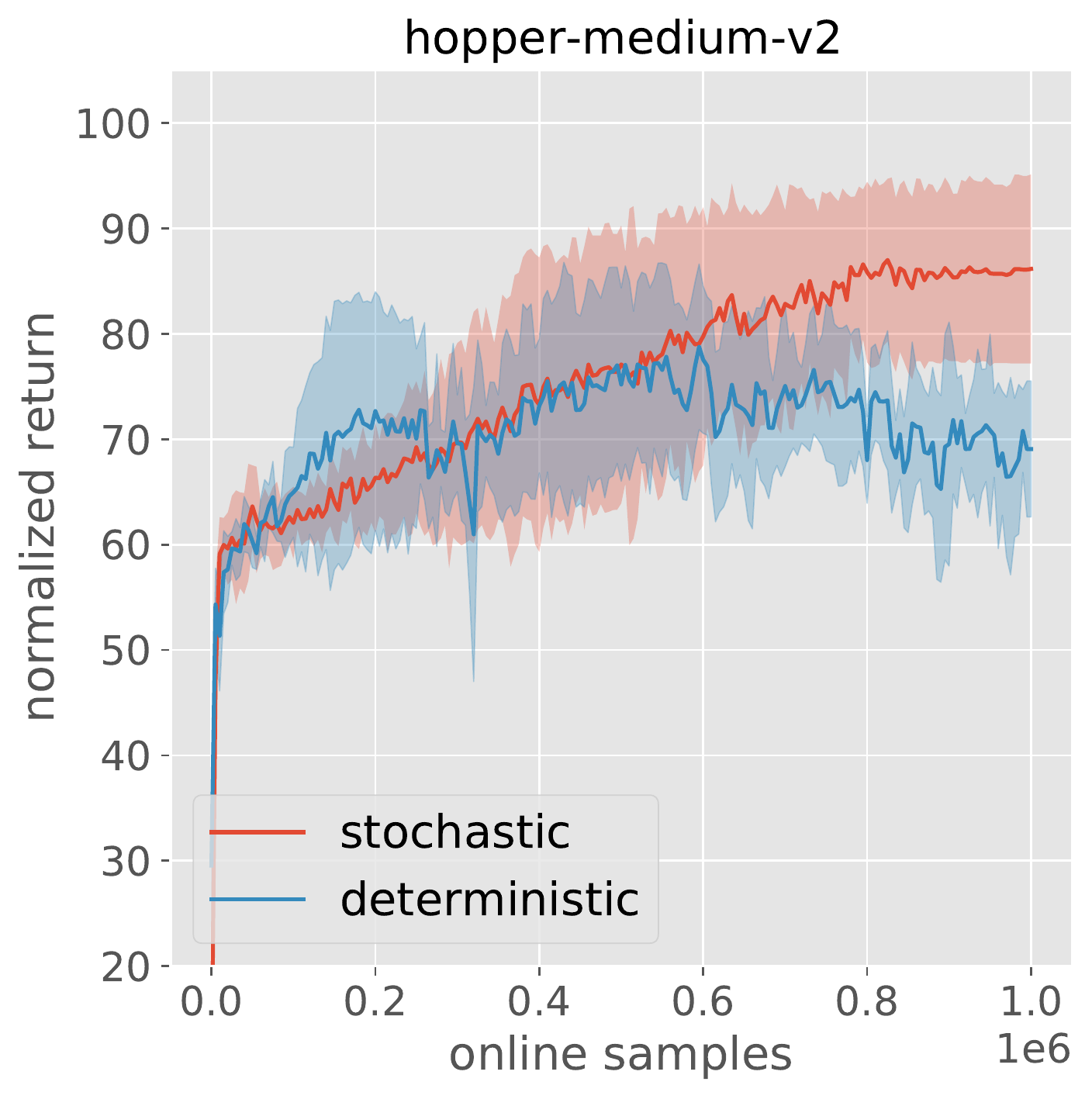}
    \includegraphics[width=0.48\columnwidth]{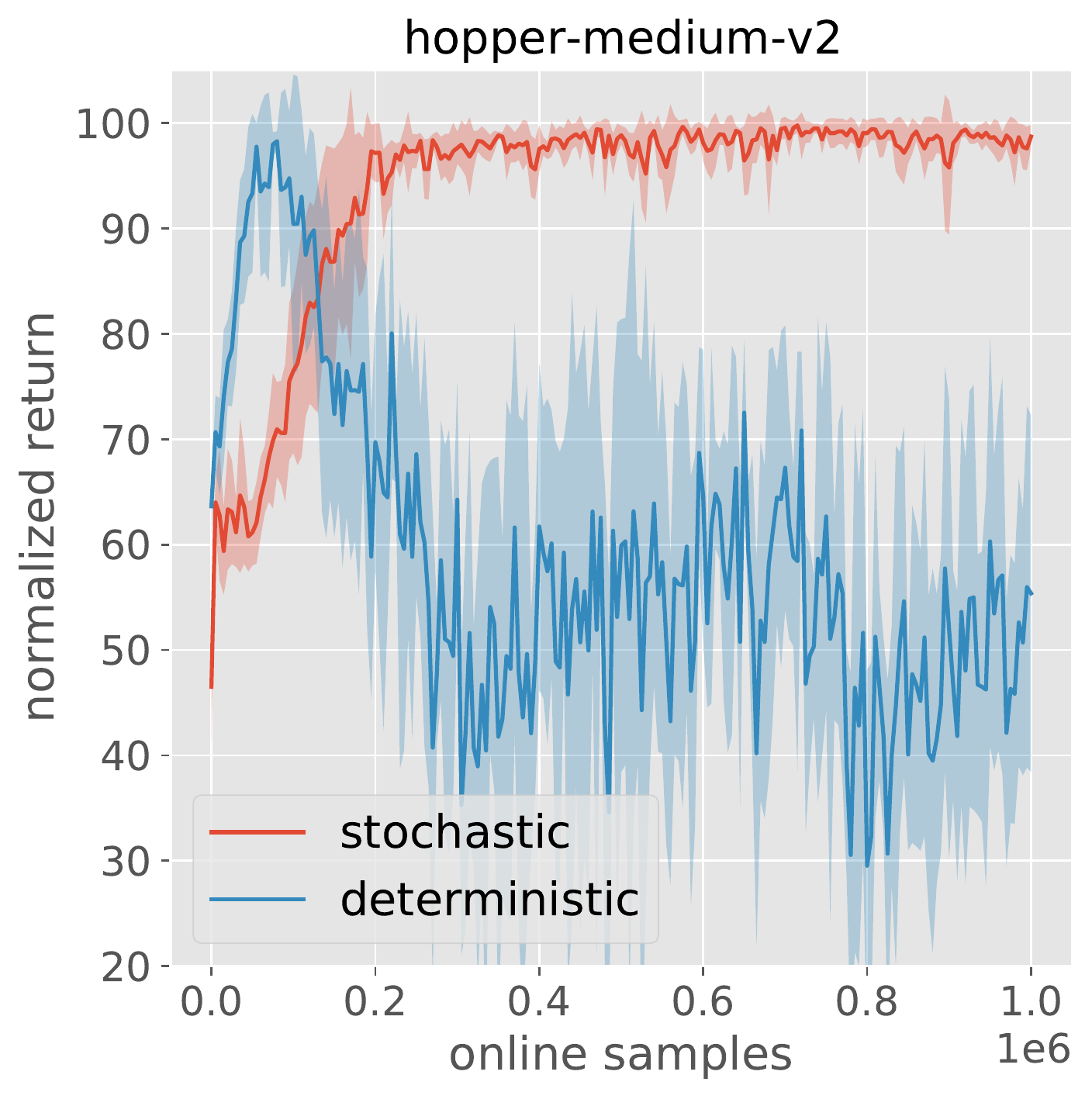}
    %%% For Arxiv
    % \includegraphics[width=0.35\columnwidth]{figure/ablation_stoch_vs_det_small.pdf}
    % \includegraphics[width=0.35\columnwidth]{figure/ablation_stoch_vs_det_large.pdf}
    \caption{Comparison of \ouracronym (red) with a deterministic variant (blue)
    in terms of training stability on the \hopper benchmark. For both small (L)
    and large (R) architectures, \ouracronym is stable whereas the performance of the deterministic
    policy declines and exhibits high variability.
    }
    \label{fig:ablation_stoch_vs_det}
\end{figure}

\paragraph{Stochastic vs Deterministic Policy.}
Stochasticity is an important component for \ouracronym to enable exploration and stable online training.
To demonstrate its effect,
we compare \ouracronym to a deterministic variant using the same finetuning framework presented in Section~\ref{sec:odt}. The deterministic variant uses the same architecture as DT, which predicts actions using a fully connected layer at the end and is optimized via the $\ell_2$ loss. Figure~\ref{fig:ablation_stoch_vs_det} compares the average performance of 5 training instances on two different model architectures. The left panel plots the results on models with small capacity where neither of them solves the environment. The performance of the deterministic variant starts to decrease after collecting $600$k online samples.
In contrast, \ouracronym is stable and gives higher returns. If we drastically increase the model capacity, as shown in the right panel, both approaches will improve but the performance of deterministic variant quickly degrades
and fluctuates, whereas \ouracronym is stable and performs consistently.
We also observe that finetuning methods are generally more stable than purely online counterparts, see Appendix~\ref{app:stability} for further details.

\begin{figure}[H]
    \centering
    \includegraphics[width=0.45\columnwidth]{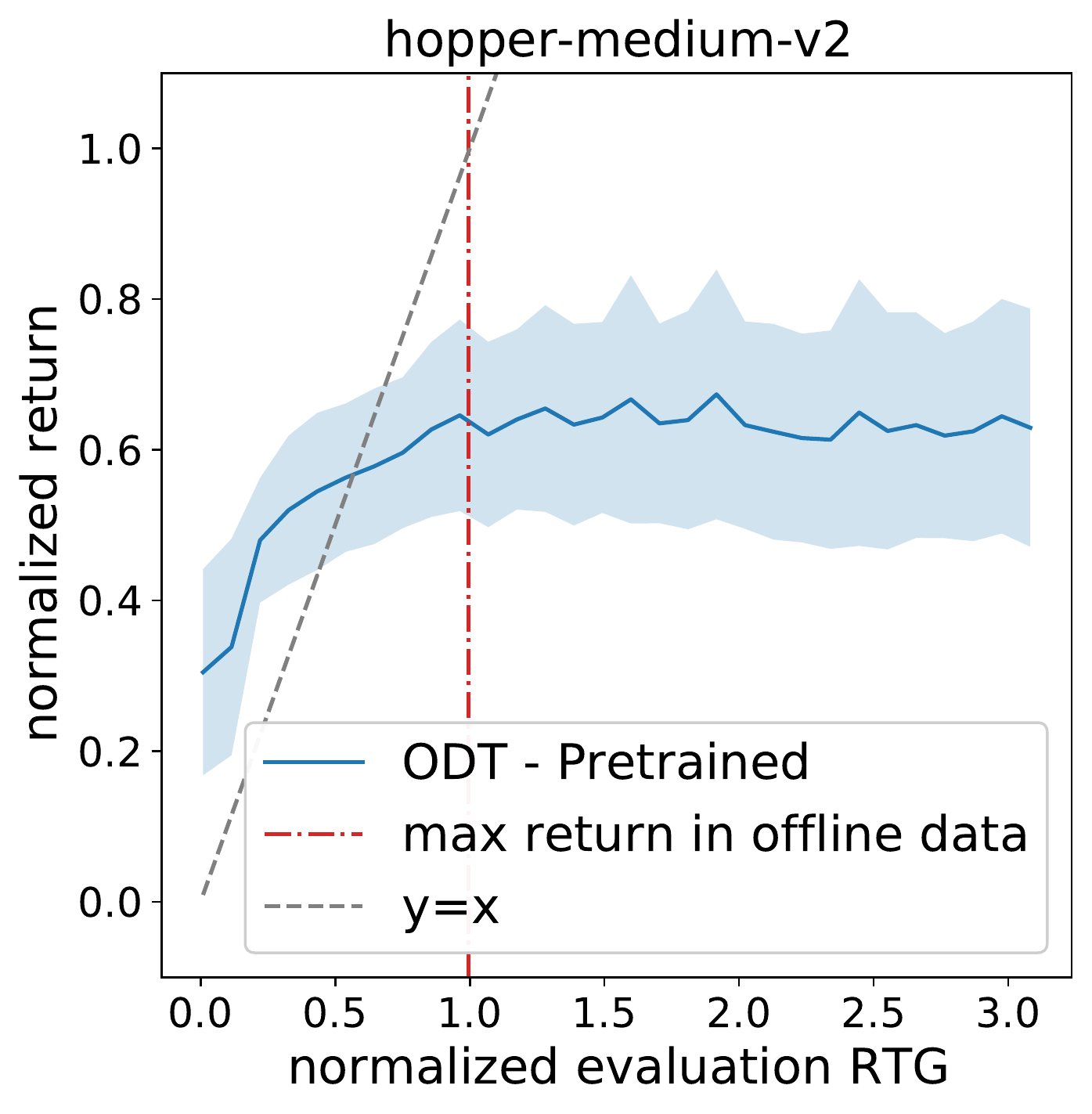}
    \includegraphics[width=0.45\columnwidth]{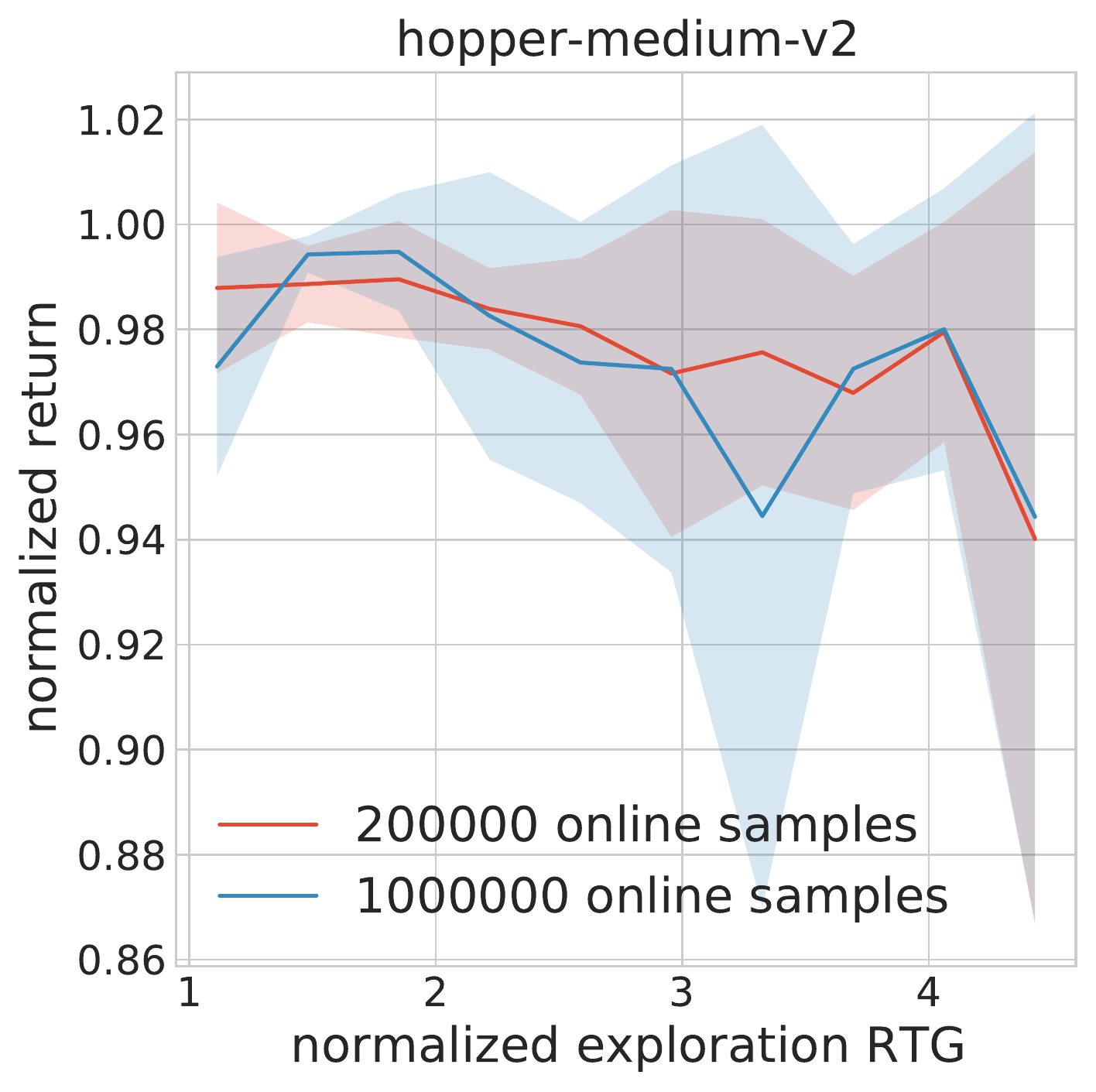}
    %%% For Arxiv
    % \includegraphics[width=0.35\columnwidth]{figure/ablation_rtg_hopper_medium_pretrain.pdf}
    % \includegraphics[width=0.35\columnwidth]{figure/hopper_med_varying_online_rtg.pdf}
    \caption{The evaluation return obtained by ODT (L) after pretraining when conditioned on various values of $g_\text{eval}$, (R) after finetuning with varying values of $g_\text{online}$ and fixed $g_\text{eval}=3600$. %\qq{title of left plot}
    }
    \label{fig:ablation_rtg}
\end{figure}

\paragraph{RTG Conditioning.}
As ODT generates return conditioned policies, we examine strategies for specifying the initial RTG token for online exploration and evaluation.
We take an offline pretrained ODT model for \hopper using \medium dataset. We vary the initial RTG token $g_\text{eval}$ for evaluation rollouts and compute the return averaged over $100$ evaluation trajectories in the left panel of
Figure~\ref{fig:ablation_rtg}. Similar to the observation made by \citet{chen2021decision}, the actual return of the offline pretrained ODT is strongly correlated with $g_\text{eval}$, and it saturates at best possible performance even when conditioned on large out-of-distribution RTGs. We built on this observation to study two mechanisms for choosing the initial RTG token $g_\text{online}$ during online exploration.

\emph{Fixed, Scaled Expert RTG.} Figure~\ref{fig:ablation_rtg} suggests that for an offline model, if $g_\text{eval}$ is large enough, increasing its value further does not change the return significantly. This motivates us to use a large RTG value even for online exploration, going much beyond the maximum achievable returns for the environment. This is applicable to practical situations where we have no knowledge of the expert performance.
% One natural question to ask is whether this value can be arbitrarily large. To understand that,
We examine the returns of ODT with varying values of $g_\text{online}$. Figure~\ref{fig:ablation_rtg} reports the results in its right panel. The returns are stable when $g_\text{online}$ is set to $1-2$x of the expert performance but slightly decrease afterwards along with increasing variance. This suggests that one can use high values of $g_\text{online}$ but not exceptionally large. For all of our experiments, we set $g_\text{online}$ to twice the expert performance. We note that setting $g_\text{online}$ as the expert performance results in comparable returns, see Appendix~\ref{sec:appendix_rtg_comparison}.

\begin{figure}[H]
    \centering
    \includegraphics[width=0.45\columnwidth]{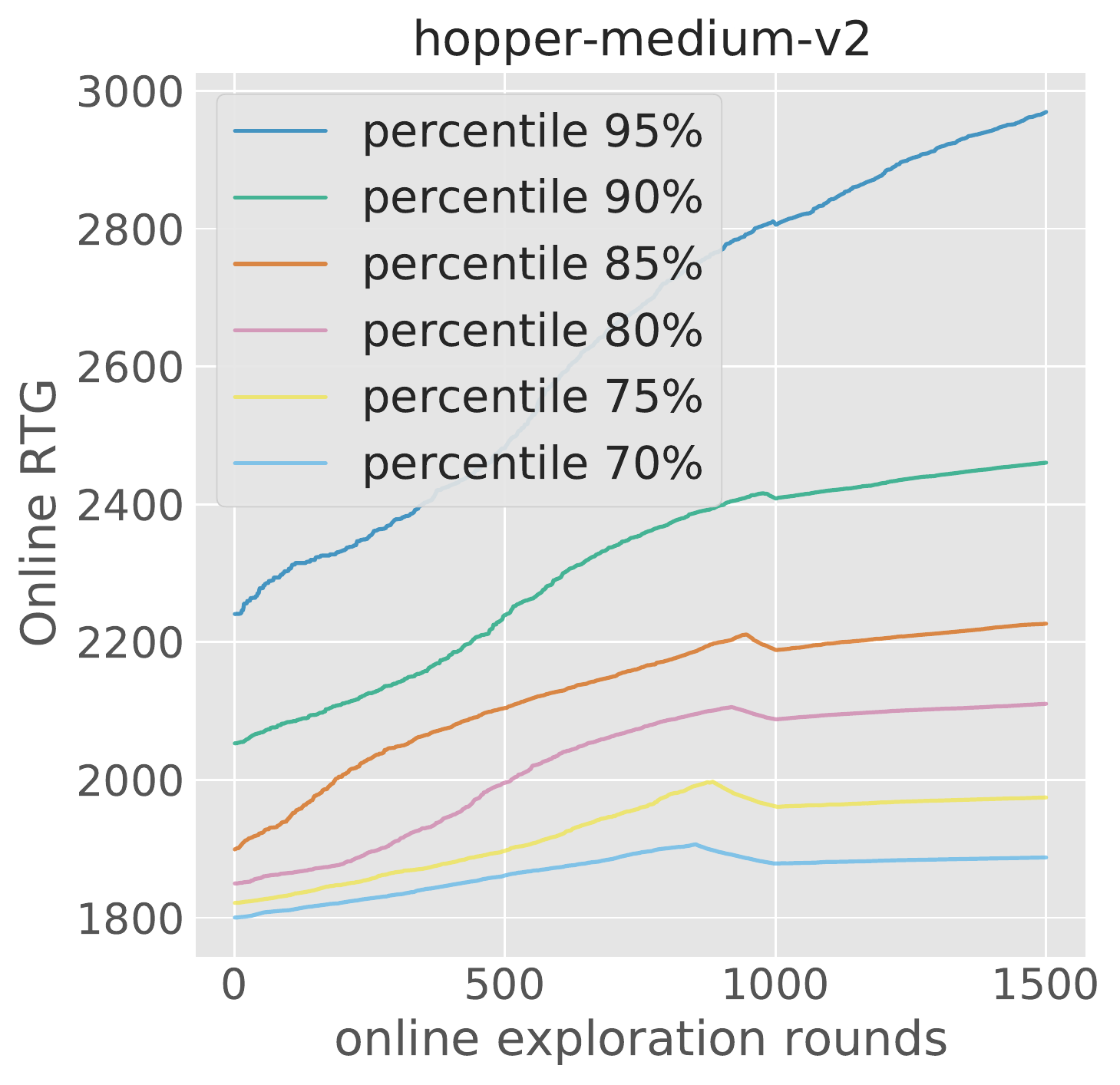}
    \includegraphics[width=0.45\columnwidth]{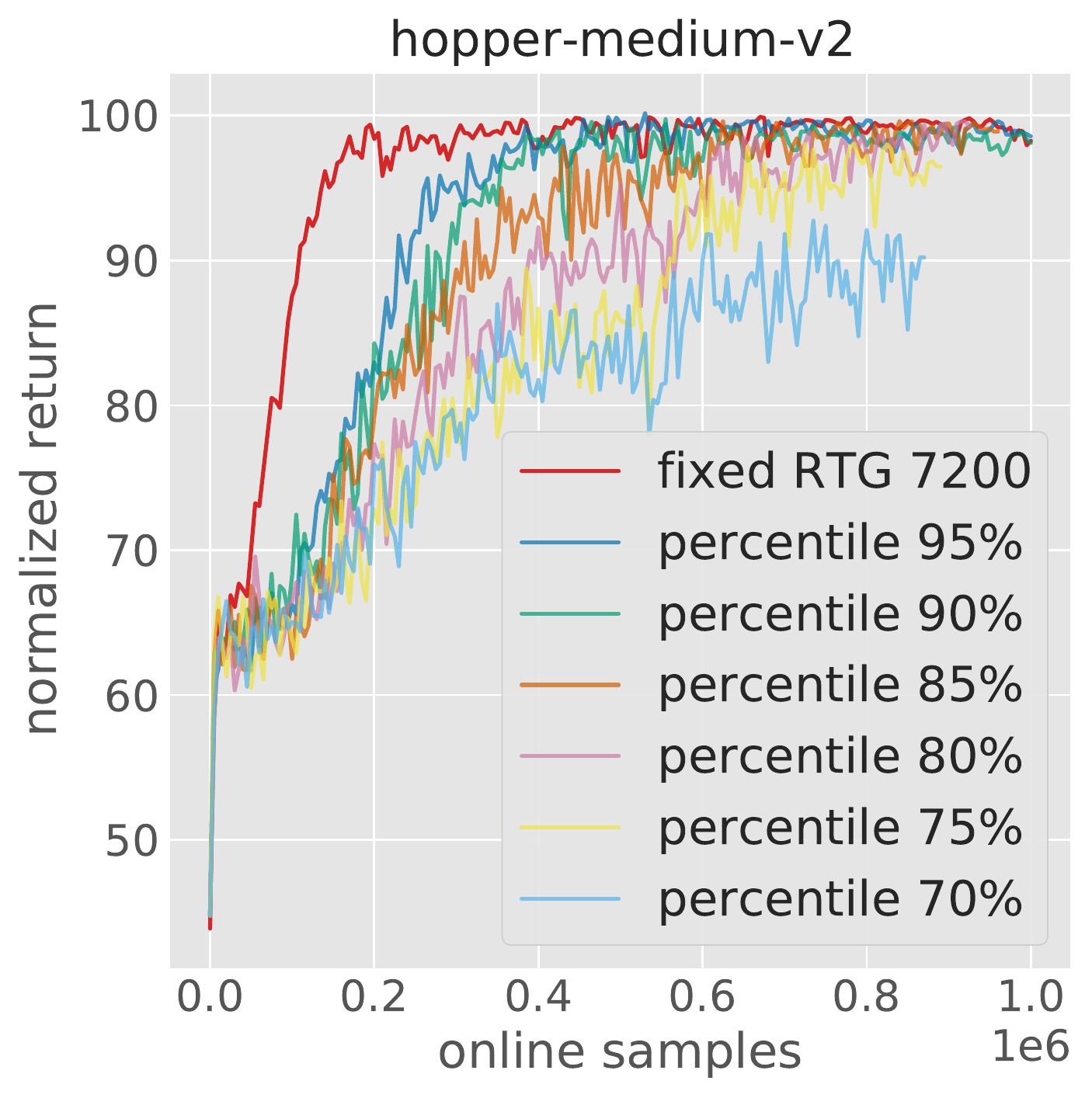}
    %%% For Arxiv
    % \includegraphics[width=0.35\columnwidth]{figure/ablation_curriculum_rtg.pdf}
    % \includegraphics[width=0.35\columnwidth]{figure/ablation_curriculum.pdf}
    \caption{Results of an \ouracronym variant that uses the curriculum heuristic to setup the exploration RTG $g_\text{online}$. (L) $g_\text{online}$ for various values of $q$. (R) Evaluation return comparison. ODT with fixed $g_\text{online} = 7200$ (red) outperforms the others, and larger $q$ leads to better performance. Besides, the runs with smaller $q$ collect fewer online samples,  this is because the return is highly correlated with the length for dense rewarded environments, see Appendix~\ref{sec:appendix_sampling}.}
    \label{fig:ablation_hopper_curriculum}
\end{figure}

\emph{Curriculum RTGs.} Another heuristic we have considered stems from curriculum learning \cite{bengio2009curriculum, bengio2015scheduled, florensa2018automatic,du2022takes}. Here, we let $g_\text{online}$ be the non-stationary $q$-th percentile of the returns of the trajectories stored in the replay buffer. Figure~\ref{fig:ablation_hopper_curriculum} reports the results. The fixed, scaled RTG strategy outperforms this heuristic.

\begin{figure}[H]
    \centering
    \includegraphics[width=0.9\columnwidth]{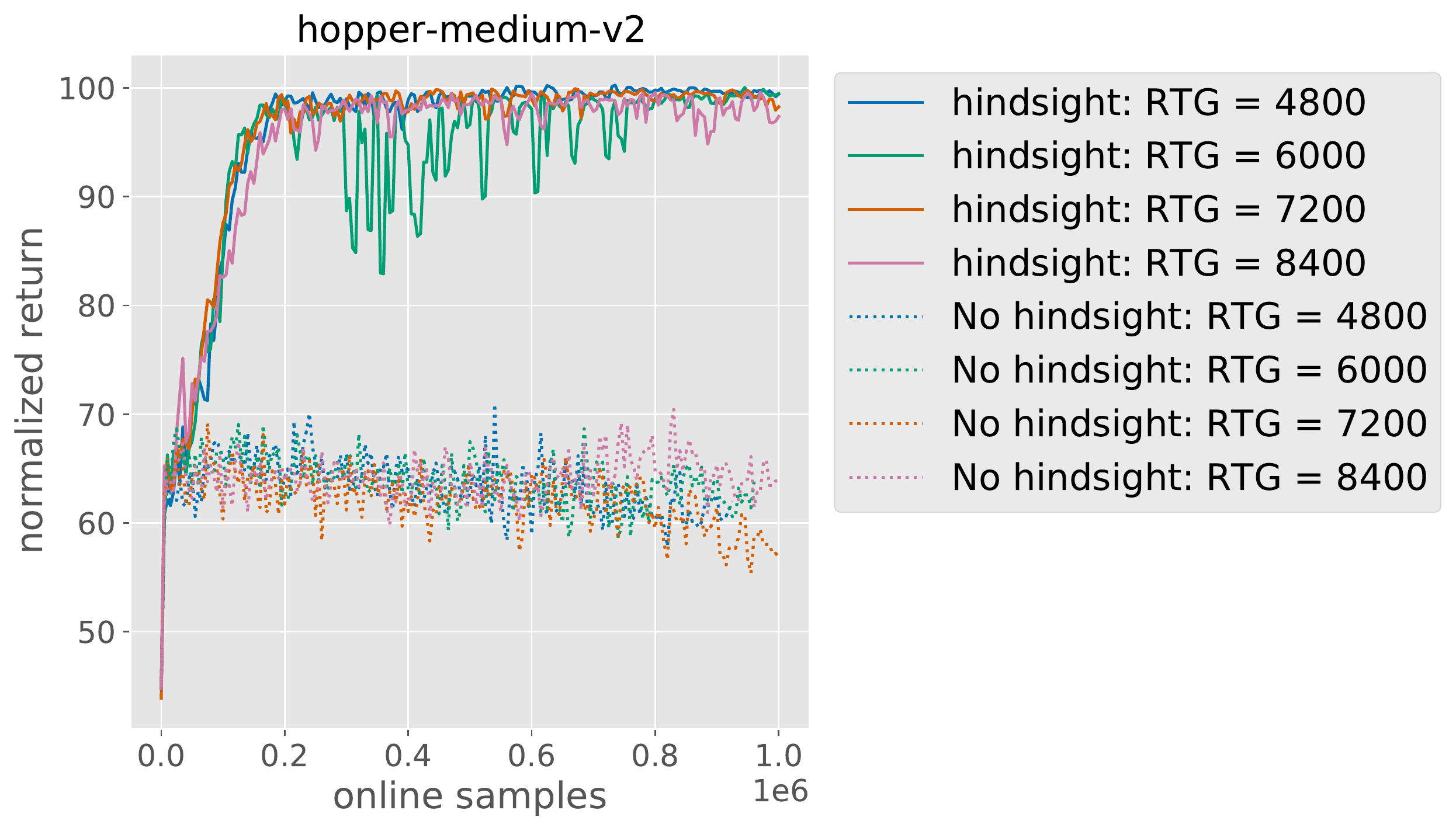}
    %%% For Arxiv
    % \includegraphics[height=0.35\columnwidth]{figure/ablation_hindsight_hopper_medium.pdf}
    \caption{Comparison of ODTs trained with hindsight return relabeling enabled or disabled on \hopper using \medium dataset. All the other hyperparameters are the same for both methods. Results are averaged over $5$ training instances.}
    \label{fig:ablation_hindsight}
\end{figure}

\paragraph{Hindsight Return Relabeling.} Figure~\ref{fig:ablation_hindsight} inspects the return of ODT with and without hindsight return relabeling (Line~\ref{alg:odt_hindsight} of Algorithm~\ref{alg:dt-train}). In the absence of relabeling, ODT quickly saturates to suboptimal performance and thus, confirming the importance of fixing the RTG tokens prior to appending the trajectory to the training batch of \ouracronym.

\section{Discussion}
In recent years, significant progress has been made in policy optimization via deep RL algorithms,
including policy gradients \cite{levine2013guided, schulman2015trust, lillicrap2015continuous, schulman2017proximal}, Q-learning \cite{lange2010deep, mnih2015human, van2016deep, gu2016continuous, wang2016dueling}, actor-critic methods \cite{schulman2015high, mnih2016asynchronous, lillicrap2015continuous, gu2016q, sac1}, and model-based learning~\cite{kaiser2019model,zhang2019solar,lu2020reset}.
Decision Transformer (DT) \cite{chen2021decision}, and the closely related work by \citet{janner2021offline}, provide an alternate perspective by framing offline RL as a sequence modeling problem and solving it via techniques from supervised learning.
% Due to its algorithmic simplicity,
This provides a simple and scalable framework, including extensions to multi-agent RL \cite{meng2021offline}, transfer learning \cite{boustati2021transfer}, and richer forms of conditioning \cite{puttermanpretraining,furuta2021generalized}.

% One limitation of DT is its offline nature.
 We proposed \ouracronym, a simple and robust algorithm for finetuning a pretrained DT in an online setting, thus further expanding its scope to practical scenarios with a mixture of offline and online interaction data.
In the future, it would be interesting to investigate whether a supervised learning approach can account for purely online RL.
% Our work presents empirical evidences that supervised learning can provide sample-efficient solutions to RL problems in certain settings. For future work, we would like to investigate whether such approach accounts for pure online RL.
Our experiment results suggest that \ouracronym is preferred by some but not all the environments.
It would be interesting to probe the limits of supervised learning algorithms for RL more broadly, in a similar spirit to \citet{emmons2021rvs}.
For example,
% There are also research attentions that studies the limitations of algorithms for RL.
% \ouracronym can be considered as a behavior cloning (BC) algorithm.
\citet{kumar2021should} characterize scenarios where RL algorithms outperform classic BC algorithms for offline RL.
Similarly, \citet{ortega2021shaking} point out that sequence modeling approaches for control can create delusions where the agent mistakes its own actions for task evidence, and propose to treat actions as causal interventions.
Finally, we are keen to develop notions of generalization~\cite{kirk2021survey,grover2018evaluating} for ODT and related RL frameworks inspired by supervised learning.
We believe these works serve as useful guidance for future work.
% It would be interesting to probe the fundamental properties of environments and algorithms that greatly influence ODT's performance.

\subsection*{Acknowledgement}
The authors thank Brandon Amos, Olivier Delalleau, Maryam Fazel-Zarandi, Jiatao Gu, Ilya Kostrikov, Kevin Lu, Mike Rabbat, Shagun Sodhani, Yuandong Tian, Mary Williamson, Lin Xiao
and Denis Yarats for insightful discussions.

% \input{dt_vs_sac}

% \clearpage
\bibliography{main}
\bibliographystyle{icml2022}

\clearpage
\appendix
\onecolumn

\section{Additional Ablation Study}
\label{sec:appendix_additioanl_ablation}

\subsection{Replay Buffer Initialization}
\label{sec:appendix_buffer_init}
\begin{figure}[H]
     \centering
     \includegraphics[width=0.55\columnwidth]{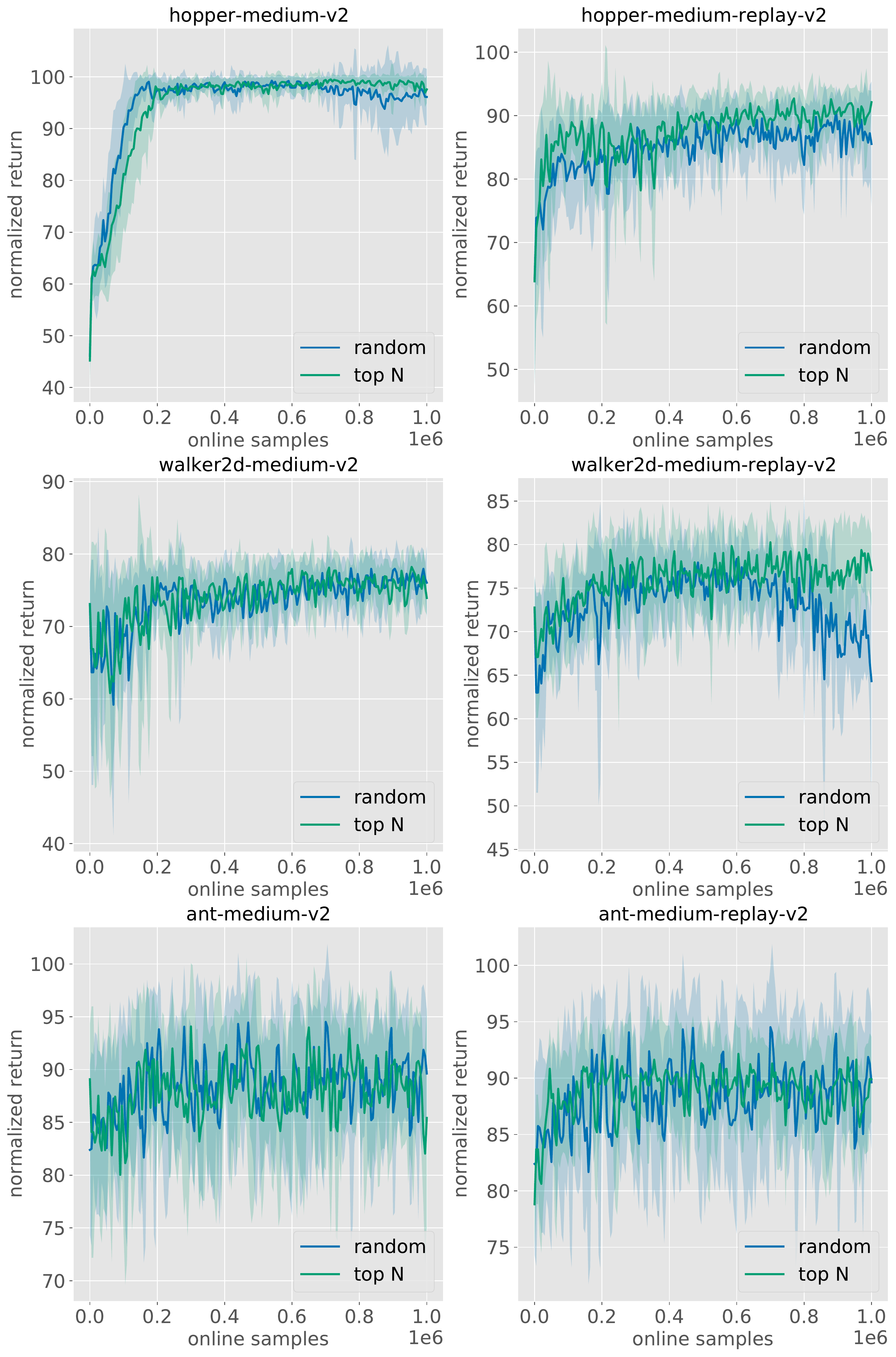}
     %%% For Arxiv
     % \includegraphics[width=0.6\columnwidth]{figure/ablation_buffer_init.pdf}
     \caption{Comparison of two strategies for initializing the replay buffer $\buffer$: (1) use the top trajectories with highest returns in the offline dataset $\offlinedata$, and (2) randomly select trajectories from $\offlinedata$. The size of $\buffer$ is 1000. For each environment,  all the other hyperparameters for both methods are the same as in Section~\ref{sec:appendix_hp}. Results are averaged over $10$ training instances with different seeds. }
     \label{fig:ablation_buffer_init}
 \end{figure}

We initialize the replay buffer $\buffer$ by the top trajectories with highest returns in the offline dataset $\offlinedata$, see Line~\ref{alg:odt_buffer_init} in Algorithm~\ref{alg:odt}. Another natural initialization strategy is to randomly select a subset of $N$
trajectories in $\offlinedata$. Figure~\ref{fig:ablation_buffer_init} compares these two strategies in multiple environments. The top $N$
strategy slightly outperforms the random initialization strategy.

\section{Additional Design Choices for ODT}
\label{sec:appendix_design}
We here discuss the following two configurable components that are critical for the actual performance of ODT. Both of them have an impact on how ODT models the long-horizon dependency for its policy. Experiments demonstrating their influences are also presented below.

\begin{figure}[H]
    \centering
    \includegraphics[width=0.9\columnwidth]{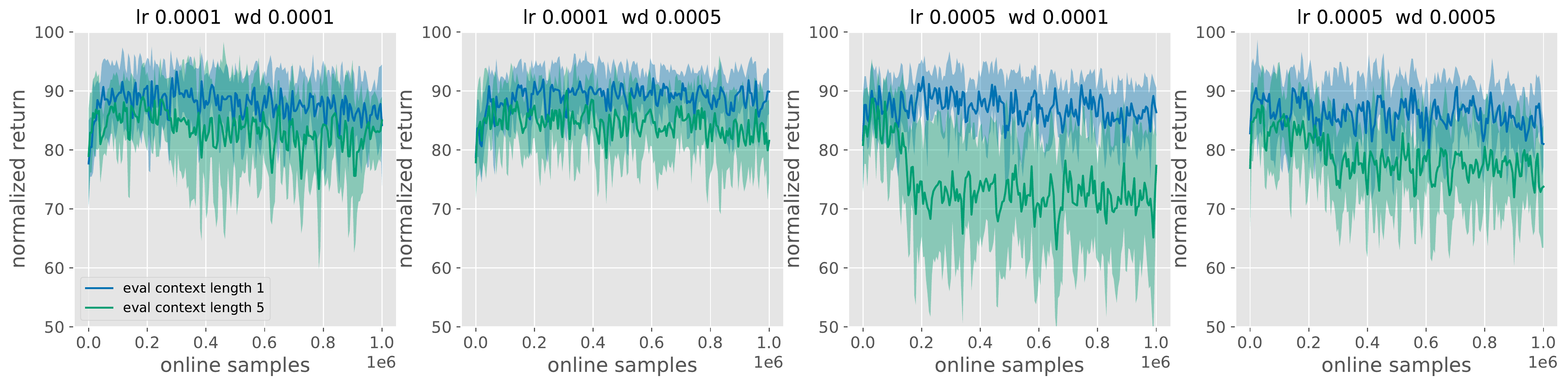}
    \caption{Comparison of the evaluation context length for finetuning ODT for \ant with \medreplay offline dataset. For varying values of learning rates and weight decays, context length $1$ consistently outperforms $5$. Results are averaged over 10 training instances with different seeds.
    }
    \label{fig:ablation_eval_len}
    % \vskip1em
% \end{figure}
% \begin{figure}[H]
    \centering
    \includegraphics[width=0.9\columnwidth]{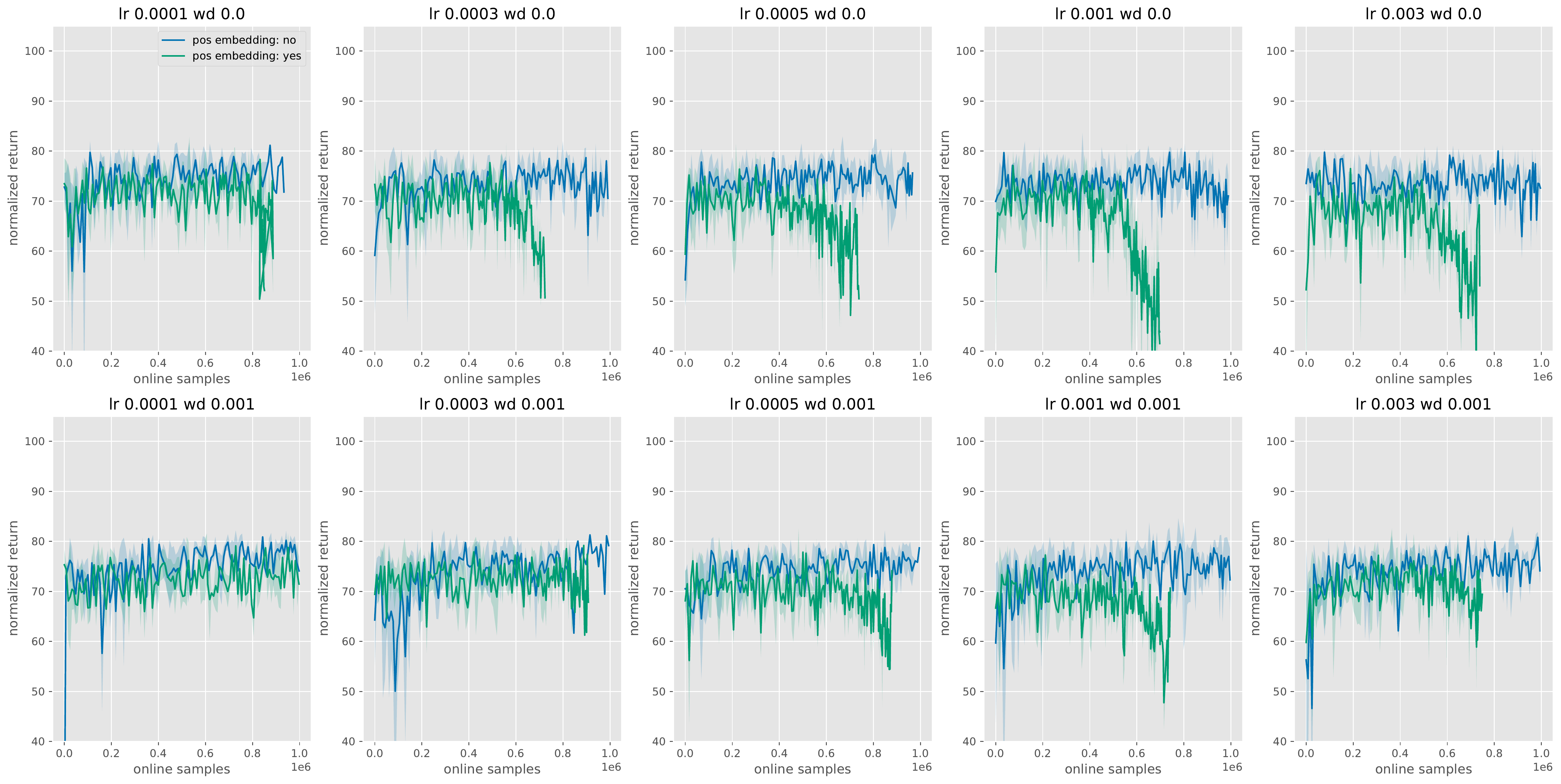}
    \caption{The online finetuning of ODT for \walker with \medium offline dataset prefers to turn off the positional embedding. For varying values of learning rates and weight decays, training without positional embedding consistently outperforms. Results are averaged over 10 training instances with different seeds.
    }
    \label{fig:ablation_pos_embedding}
\end{figure}

\paragraph{Evaluation Context Length} As mentioned in Section~\ref{sec:notation}, the context length at evaluation time is a hyperparameter we can choose. This parameter adjusts the length of past states and RTGs that the agent's present action depends on. In the edge case where the evaluation context length is $1$, the agent's future action only depends on the present state, which means the ODT policies become Markovian, and vice versa.
In our experiments, we have found that setting context length $5$ at evaluation generally leads to high return across many environments for online finetuning, but the \ant task prefers context length $1$, see Figure~\ref{fig:ablation_eval_len}. However, the preferences are mixed for offline training. We summarize the best hyperparameters we have found in Section~\ref{sec:appendix_hp}.
% \begin{figure}[H]
%     \centering
%     \includegraphics[width=\columnwidth]{figure/ablation_eval_len_ant_replay.pdf}
%     \caption{Comparison of the evaluation context length for finetuning ODT for \ant with \medreplay offline dataset. For varying values of learning rates and weight decays, context length $1$ consistently outperforms $5$. Results are averaged over 10 training instances with different seeds.
%     }
%     \label{fig:ablation_eval_len}
%     % \vskip1em
% \end{figure}

\paragraph{Positional Embedding} In language modeling, positional embeddings are used to equip the input words with their positional information. DT uses the absolute positional embedding which is trained to represent the timesteps of a trajectory, and is added to the embedding of states, actions, and RTGs to help them determine their exact positions within a trajectory. Alternatively, one can use the relative positional embeddings which only account for the order of the states in the input sub-trajectory. However, for problems with non-negative dense rewards, the RTG sequence is monotonically decreasing so that the positional information is partially contained.
We have found for some \gym tasks in the D4RL benchmark, removing positional embedding improves ODT's finetuning performance, see Figure~\ref{fig:ablation_pos_embedding} for an example. We conjecture this is due to the fact that the representations of states, actions and RTGs are more disentangled in this case. On the other hand, for goal-reaching tasks with sparse rewards e.g. \antmaze, the RTG sequence does not contain timestep information, and positional embedding is needed to stitch the sub-trajectories. Throughout our finetuning experiments in Section~\ref{sec:expr_comp}, we remove the positional embedding for all the \gym tasks but use them for the \antmaze tasks.

\section{Hyperparameters of ODT}
\label{sec:appendix_hp}

We summarize the architecture and other design hyperparameters that we found to work best for ODT in this section. Worthy of note, we use a model with much larger capacity than DT. The transformer we use consists of 4 layers with 4 attention heads, and the embedding dimension is 512. Our intuition is that ODT has more complex structure including the stochasticity to enables online adaption, for which optimization becomes more challenging. Increasing the model capacity might help alleviate the difficulty. Meanwhile, the hyperparameters that result in high return in offline training and online finetuning might not be the same, and we report them separately below. 

For all the experiments, we optimize the policy parameter $\theta$ by the LAMB optimizer~\cite{you2019large}, for which we report the learning rates and weight decay parameters below. To optimize the dual parameter $\lambda$ in problem~\eqref{eq:sdt_opt_lambda}, we optimize the transformed parameter $\log \lambda$ by the Adam optimzier~\cite{kingma2014adam} with fixed learning rate $0.0001$.

As an additional note, the v2 \antmaze datasets we use for our experiments have updated timeout information and we obtain them from the D4RL group. For more results, visit \url{https://sites.google.com/view/onlinedt/home}.
\subsection{Offline Pretraining}
\begin{table}[H]
    \centering
    \small
    \begin{tabular}{l    l  }
    \toprule
         hyperparameter & value \\ \midrule
         Number of layers & $4$ \\
         Number of attention heads & $4$\\
         Embedding dimension & $512$\\
         Training Context Length $K$ & $20$\\
         Dropout & $0.1$\\
         Nonlinearity function & ReLU\\
         Batch size & $256$\\
         Learning rate & $0.0001$ \\
         Weight decay & $0.001$ \\
         Gradient norm clip & $0.25$ \\
         Learning rate warmup & linear warmup for $10^4$ training steps \\
         Target entropy $\beta$ & $-\text{dim}(\mathcal{A})$ \\
         Total number of updates & $10^5$ \\ \bottomrule
    \end{tabular}
    \caption{The common hyperparameters that are used to pretrain \ouracronym.}
    \label{tbl:odt_common_offline}
    \vskip3em
    \begin{tabular}{l || c  c  c  c  c  c  c  c  c}
    \toprule
         dataset    &     eval context length & positional embedding & $g_\text{eval}$ \\ \midrule
        hopper-medium & $20$ & yes & $3600$ \\
        hopper-medium-replay & $20$ & yes & $3600$ \\
        walker2d-medium & $1$ & no & $5000$\\
        walker2d-medium-replay & $5$ & no & $5000$\\
        halfcheetah-medium & $5$ & no & $6000$ \\
        halfcheetah-medium-replay & $1$ & no & $6000$ \\
        ant-medium & $5$ & yes & $6000$ \\
        ant-medium-replay & $20$ & yes & $6000$ \\ \midrule
        antmaze-umaze & $1$ & yes & $1$ \\
        antmaze-umaze-diverse & $1$ & yes & $1$ \\ \bottomrule
    \end{tabular}
    \caption{The hyperparameters that we found to work best to pretrain \ouracronym in each domain.}
    \label{tbl:odt_hp_offline}
\end{table}
For offline training, all the tasks we consider share most of the hyperparameters except evaluation context length and positional embedding. Following~\cite{sac1, sac2}, we set the target entropy $\beta$ to be the negative action dimensions.
Table~\ref{tbl:odt_common_offline} lists the common hyperparameters and Table~\ref{tbl:odt_hp_offline} lists the domain specific ones. Table~\ref{tbl:odt_hp_offline} also reports the RTG $g_\text{eval}$ we use for evaluation rollouts.

\subsection{Online Finetuning}
Likewise, we report the shared hyperparameters in Table~\ref{tbl:odt_common_online} and domain specific ones in  Table~\ref{tbl:odt_hp_online}. Note that we have found embedding dimension $1024$ performs much better than $512$ for antmaze-umaze-diverse in finetuning, yet worse in offline training. One of our important finding is that pretraining till convergence might hurt the exploration performance, and we use much fewer number of pretraining updates than offline models in this scenario.

\begin{table}[h]
    \centering
    \small
    \begin{tabular}{l  l  }
    \toprule
         hyperparameter & value \\ \midrule
         Number of layers & $4$ \\
         Number of attention heads & $4$\\
         Embedding dimension & $1024$ for antmaze-umaze-diverse and $512$ for all the other tasks \\
         Training Context Length $K$ & $20$\\
         Dropout & $0.1$\\
         Nonlinearity function & ReLU\\
         Batch size & $256$\\
         Updates between rollouts & $300$ \\
         Gradient norm clip & $0.25$ \\
         Learning rate warmup & linear warmup for $10^4$ training steps \\
         Target entropy $\beta$ & $-\text{dim}(\mathcal{A})$ \\ \bottomrule
    \end{tabular}
    \caption{The common hyperparameters that are used to finetune \ouracronym.}
    \label{tbl:odt_common_online}
\end{table}

\begin{table}[h]
    \centering
    \scriptsize
    \resizebox{\columnwidth}{!}{
    \begin{tabular}{l || c  c  c  c  c  c  c  c  c}
    \toprule
         dataset    &    \shortstack{pretraining\\updates}  & \shortstack{buffer\\size} & \shortstack{embedding\\size}&
         \shortstack{learning\\rate} & \shortstack{weight\\decay} & \shortstack{eval\\context\\length} & $g_\text{eval}$ & $g_\text{online}$ & position \\ \midrule
        hopper-medium & $5000$ & $1000$ & $512$ & $0.0001$ & $0.0005$ & $5$ & $3600$ & $7200$ & no \\
        hopper-medium-replay & $5000$ & $1000$ & $512$ & $0.002$ & $0.0001$ & $5$ & $3600$ & $7200$ & no\\
        walker2d-medium & $10000$ & $1000$ & $512$ & $0.001$ & $0.001$ & $5$ & $5000$ & $10000$ & no\\
        walker2d-medium-replay & $10000$ & $1000$ & $512$ & $0.001$ & $0.001$ & $5$ & $5000$ & $10000$ & no\\
        halfcheetah-medium & $5000$ & $1000$ & $512$ & $0.0001$ & $0.0005$ & $5$ &$6000$ &$12000$ & no\\
        halfcheetah-medium-replay & $5000$ & $1000$ & $512$ & $0.0001$ & $0.0005$ & $5$ &$6000$ &$12000$ & no\\
        ant-medium & $10000$ & $1000$ & $512$ & $0.0015$ & $0.0001$ & $1$ & $6000$ &$12000$ & no \\
        ant-medium-replay & $10000$ & $1000$ & $512$ & $0.0001$ & $0.0005$ & $1$ & $6000$ &$12000$ & no \\ \midrule
        antmaze-umaze & $7000$ & $1500$ & $512$ & $0.001$ &  $0$ & $5$ & $1$ & $2$ & yes \\
        antmaze-umaze-diverse & $7000$ & $1500$ & $1024$ & $0.001$ &  $0$ & $5$ & $1$ & $2$ & yes \\ \bottomrule
    \end{tabular}
    }
    \caption{The hyperparameters that we use to finetune \ouracronym in each domain.}
    \label{tbl:odt_hp_online}
\end{table}

\section{Comparison of the Exploration RTG $g_\text{online}$}
\label{sec:appendix_rtg_comparison}

\begin{table}[H]
    \centering
    % \scriptsize
    \small
    \begin{tabular}{l || c  c }
    \toprule
        dataset    &    $g_\text{online} = 2g^*$ & $g_\text{online} = g^*$ \\ \midrule
        hopper-medium &  97.54 & 95.70\\
        hopper-medium-replay & 88.89 & 86.09 \\
        walker2d-medium & 76.79 & 74.26\\
        walker2d-medium-replay & 76.86 & 71.76\\
        halfcheetah-medium & 42.16 & 42.97\\
        halfcheetah-medium-replay & 40.42 & 40.85 \\
        ant-medium & 90.79 &  88.94\\
        ant-medium-replay & 91.57 &  90.21\\ \midrule
        antmaze-umaze & 88.5 & 87.7\\
        antmaze-umaze-diverse & 58.19 & 59.4 \\ \bottomrule
    \end{tabular}
    \caption{The evaluation return obtained by ODT after collection $200$k online samples, using different values of exploration RTG $g_\text{online}$. The returns are comparable. Results are averaged over $10$ training instances with different seeds.}
    \label{tbl:odt_rtg_comparison}
\end{table}

Let $g^*$ denote the expert return. We train ODTs where the initial RTG token for the exploration rollouts are set to $g^*$ and $2g^*$, respectively. Table~\ref{tbl:odt_rtg_comparison} reports the return after collecting $200$k online samples for all the environments we consider, using the same hyperparameters reported in Section~\ref{sec:appendix_hp}. The obtained returns are comparable.

\section{Training Stability Comparison}\label{app:stability}
\begin{figure}[H]
    \centering
    \includegraphics[width=0.4\columnwidth]{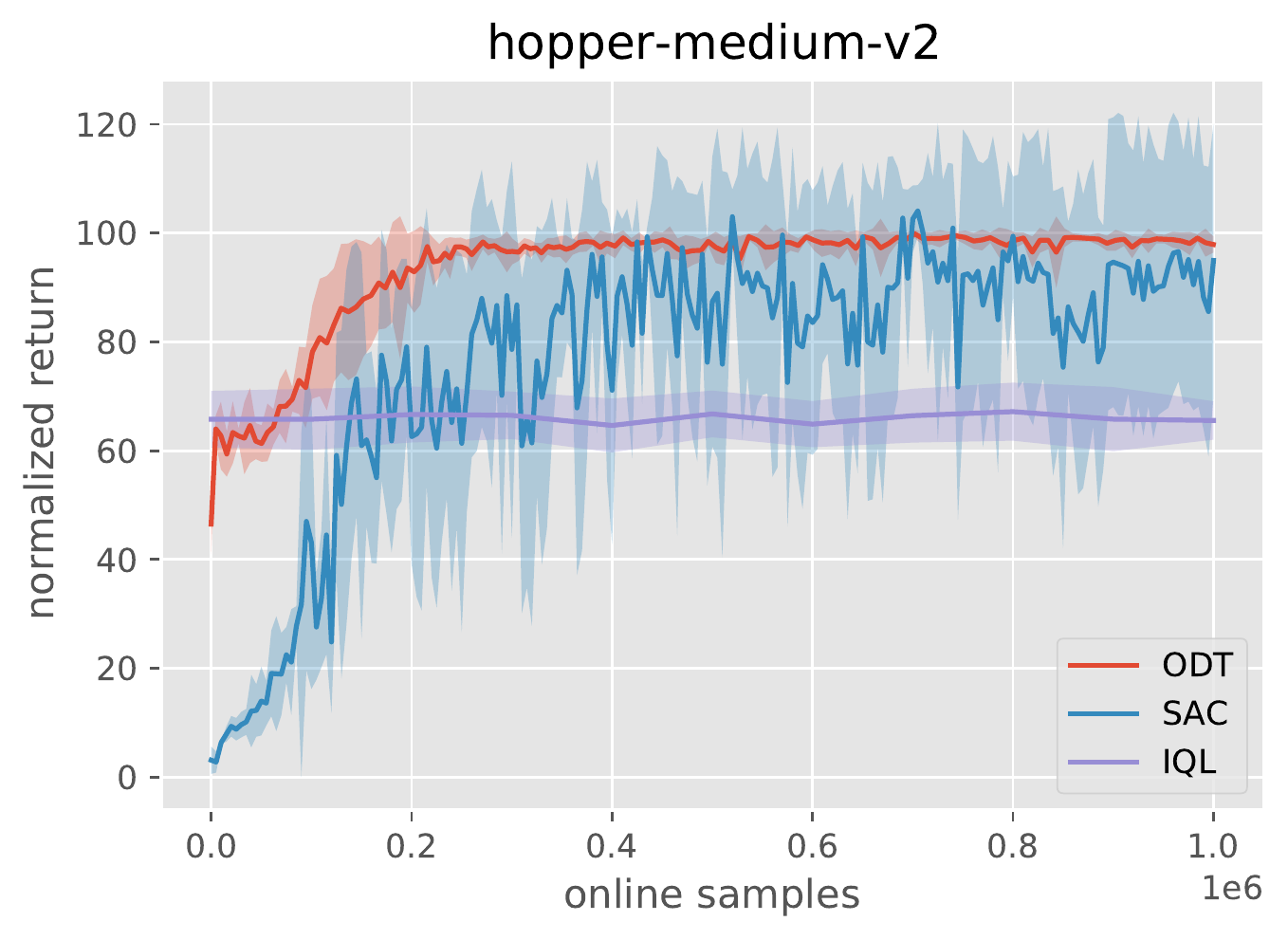}
    \caption{The training stability comparison of \ouracronym, IQL and SAC. The performance of \ouracronym improves upon pretraining using online interactions, while the performance of IQL stays almost the same as the pretraining result. Both \ouracronym and IQL are stable along the online finetuning yet SAC fluctuates. Results are averaged over 5 training instances with different seeds.}
    \label{fig:comp_stability}
\end{figure}
As noted in Section~\ref{sec:expr_ablate}, we observe that the finetuning methods are generally more stable than purely online
counterparts. Figure~\ref{fig:comp_stability} plots the results for \ouracronym, IQL and SAC on \hopper with \medium dataset. The hyperparameters are set as described in Section~\ref{sec:expr_comp} for IQL and SAC, and as described in Section~\ref{sec:appendix_hp} for \ouracronym. We can see that return of \ouracronym improves smoothly over the course of training and the return of IQL keeps stable without significant changes, whereas the return of SAC has high variety although it is increasing.

\section{Sampling Strategy}
\label{sec:appendix_sampling}
\begin{figure}[H]
    \centering
    \includegraphics[width=0.85\columnwidth]{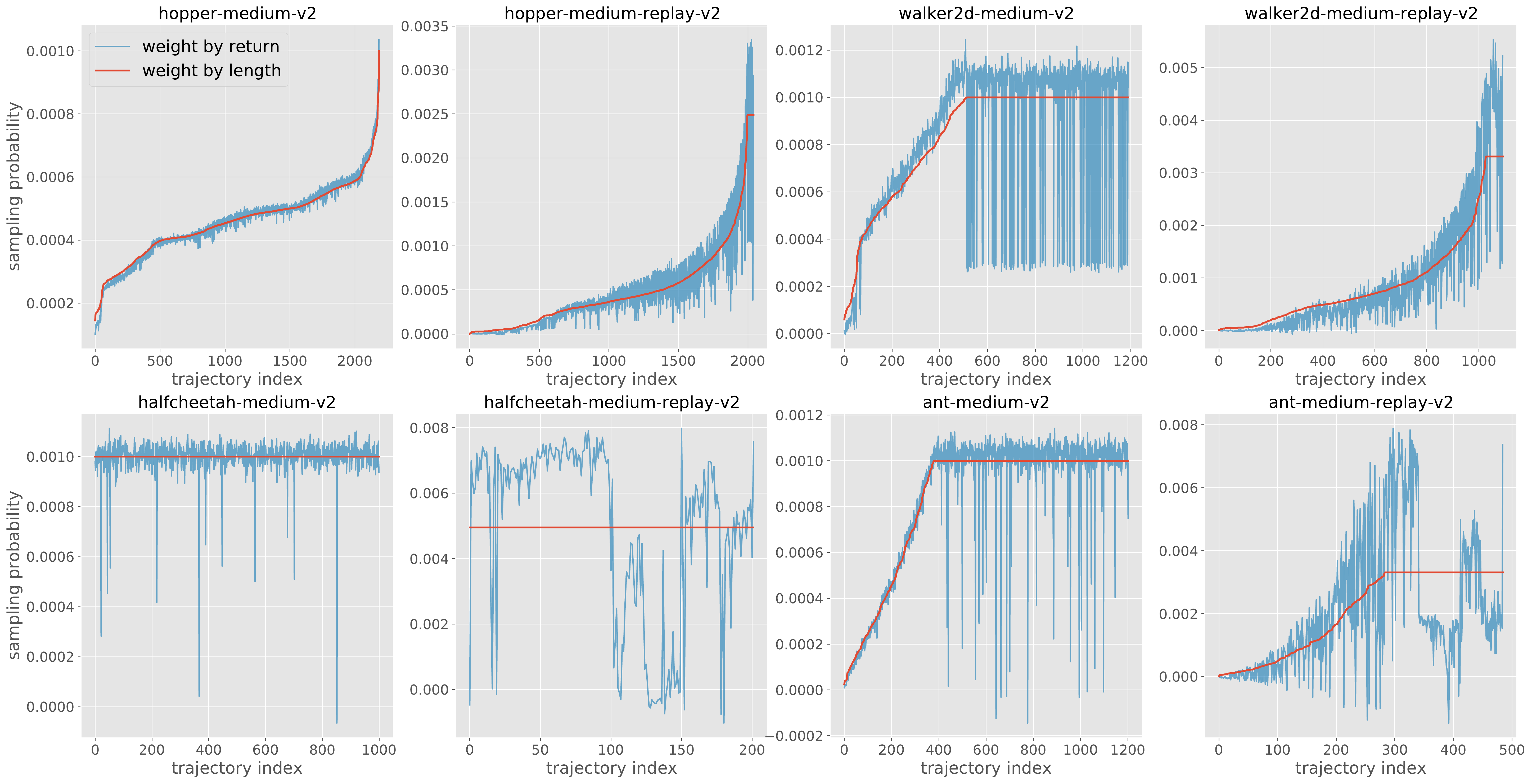}
    \caption{Comparison of our sampling strategy and importance sampling for the offline datasets collected in environments with non-negative dense rewards.}
    \label{fig:ablation_sampling}
\end{figure}

The sampling strategy presented in Algorithm~\ref{alg:dt-train} has two steps, summarized in Algorithm~\ref{alg:uniform_sampling_K}. The first step is to sample a single trajectory $\tau$ from the replay buffer $\buffer$, with probability proportional to its length:
$\P_{\buffer}(\tau) = |\tau| / \textstyle\sum_{\tau \in \buffer} |\tau|$.
Next, we uniformly sample the subsequences. Figure~\ref{fig:ablation_sampling} plots the sampling probability for each trajectory in the D4RL offline datasets for environments with non-negative dense rewards. We compare it with the importance sampling method, where the probability is proportional to the return of a trajectory. We can see that for many of the datasets, the return of a trajectory is highly correlated with its length.

\begin{algorithm}[h]
 \DontPrintSemicolon
 % \small
  \textbf{Input:} replay buffer $\buffer$, sequence length $K$.\;
  Sample a trajectory $\tau$ from $\buffer$ with probability $\P_{\buffer}(\tau) = |\tau| / \textstyle\sum_{\tau \in \buffer} |\tau|$.\;
  Compute the RTGs $g_t = \scriptstyle \sum_{t'=t}^{|\tau|} r_{t'}$, $1\leq t \leq |\tau|$. \;
  Sample a timestep $t \in {1, \ldots, |\tau|}$ uniformly at random. \;
  $\vs \leftarrow (s_t, \ldots, s_{\min(t+K-1, |\tau|)})$, $\va \leftarrow (a_t, \ldots, a_{\min(t+K-1, |\tau|)})$,
  $\vrtg \leftarrow (g_t, \ldots, g_{\min(t+K-1, |\tau|)})$.\;
  \textbf{Output:} subsequences $(\va, \vs, \vrtg)$
  \caption{Uniform Subsequence Sampling}
  \label{alg:uniform_sampling_K}
\end{algorithm}

\section{Training Dynamics: the convergence of $\lambda$}
\label{sec:appendix_lambda_convergence}
Recall that when optimizing the constrained problem~\eqref{eq:sdt_main}, we consider the Lagrangian $L(\theta, \lambda) = J(\theta) + \lambda(\beta - H^{\Tau}_\theta[\va|\vs, \vrtg])$ and update $\theta$ and $\lambda$ alternatingly. As discussed in Section~\ref{sec:odt_convergence}, under the assumption that the training converges, if $\lambda$ converges to a value in $[0, 1)$, our overall loss will converge to the desired formulation~\eqref{eq:sdt_convergence_online}.

\begin{figure}[H]
    \centering
    \includegraphics[width=0.4\columnwidth]{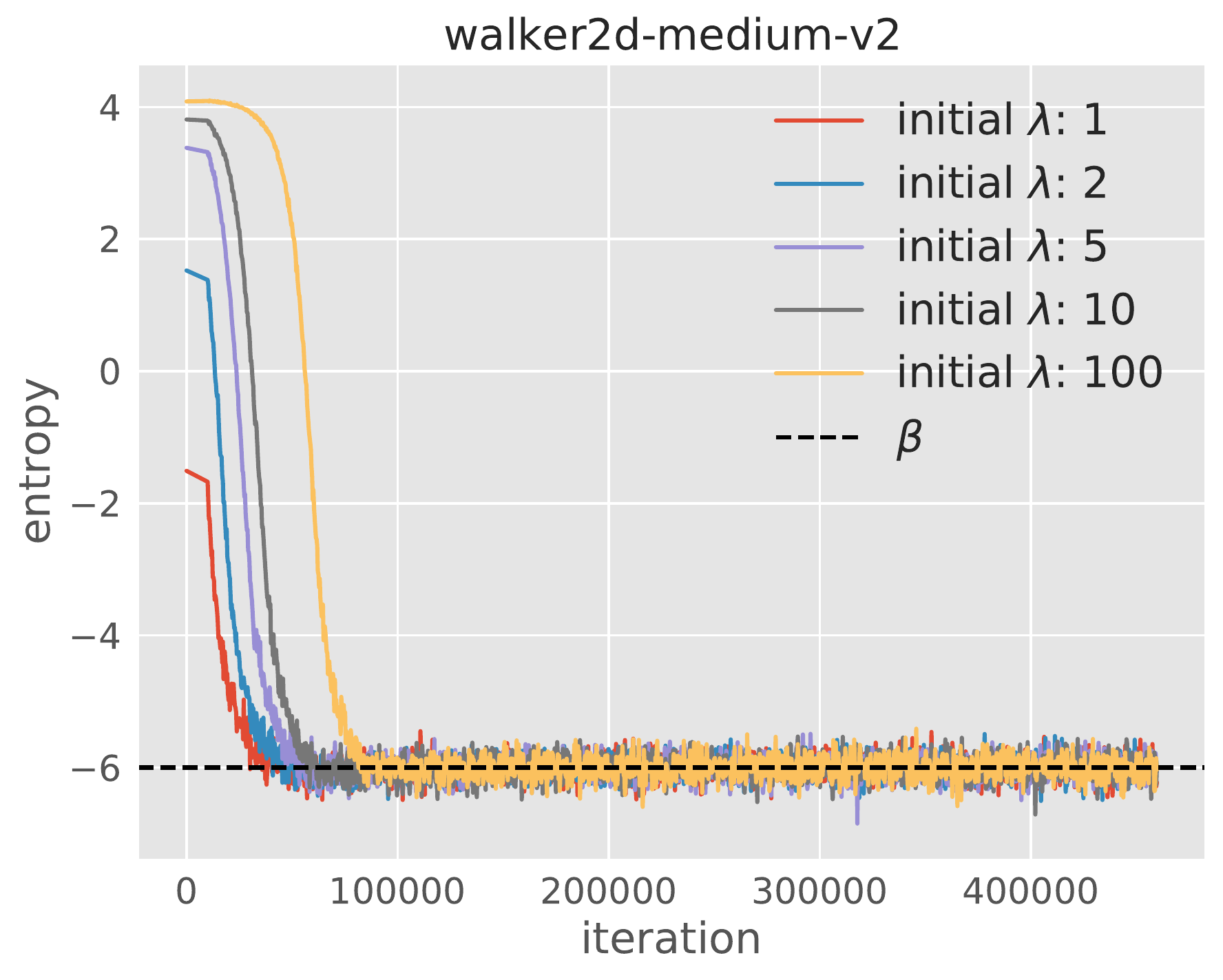}
    \includegraphics[width=0.4\columnwidth]{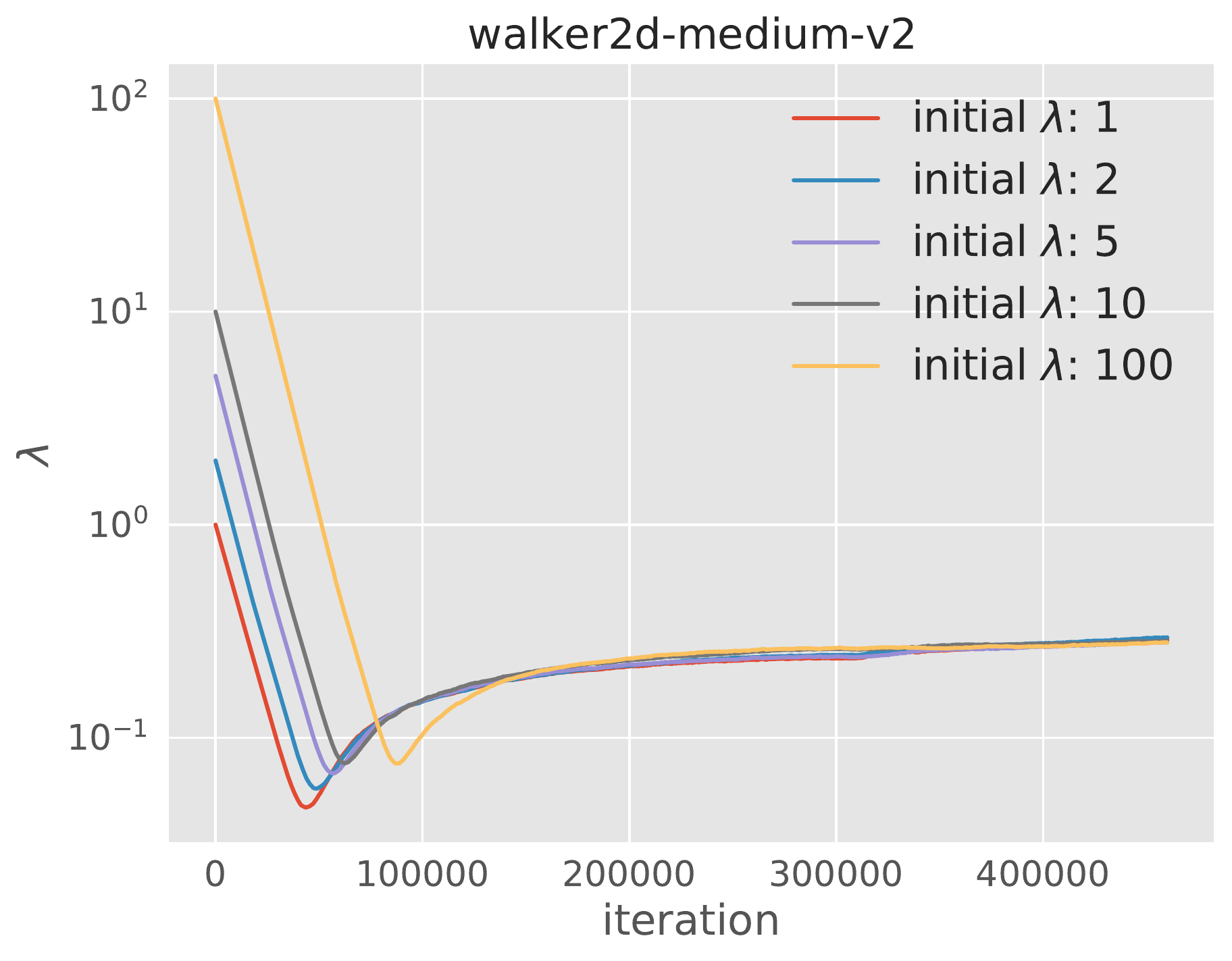}
    \caption{The estimated entropy and the dual variable $\lambda$ when optimizing problem~\eqref{eq:sdt_main}, including both pretraining and finetuning stages of ODT.}
    \label{fig:lambda_convergence}
\end{figure}

In theory, $\lambda$ can be zero if the constraint is inactive, or positive if the constraint is tight. Figure~\ref{fig:odt_convergence_example} shows an example where $\lambda$ converges to zero. Here we show another example where $\lambda$ converges to a positive value between $0$ and $1$. Figure~\ref{fig:lambda_convergence} shows five example runs for the \walker environment using \medium dataset, where we initialize $\lambda$ by $1, 2, 5, 10$, and $100$ respectively. The left panel plot the one-sample estimated entropy vs the gradient update iterations, where the first $10000$ iterations are offline pretraining, and the later ones are online finetuning. For all the five runs, we can see that the entropy converges to $\beta$, which means the constraint $H^{\Tau}_\theta[\va|\vs, \vrtg] \geq \beta$ is tight. In this scenario, $\lambda$ can be positive. The right panel plots the value of $\lambda$ over the course of training. We can see that $\lambda$ converges to a same value between $0$ and $1$ even under different initializations. 

\section{Negative results}
\label{sec:bad_heuristics}
We have tried various heuristics in a few aspects of our algorithms. For modeling, we have tried incorporating both the absolute and relative positional embedding, turning off the dropout regularization, varying the training context length $K$. For optimization strategies, we have tried the stagewise learning rate decay, the adaptive gradient updates where the number of gradient updates after each exploration rollout is the same as the newly collected trajectory. We found that these heuristics were of limited utility in preliminary experiments and hence we don't include them in the final design.

\end{document}